%% file: main.tex
\PassOptionsToPackage{table}{xcolor}
\documentclass[sigconf,authorversion,nonacm]{acmart}
\input{Section_Tex_Files/macros}

\AtBeginDocument{%
  \providecommand\BibTeX{{%
    \normalfont B\kern-0.5em{\scshape i\kern-0.25em b}\kern-0.8em\TeX}}}

\extendedtrue

\usepackage{algorithm,algpseudocode}
\usepackage{amsmath}
\usepackage{amsfonts}
\usepackage{stackengine}
\newcommand\Circle[1]{%
  \stackengine{0pt}{#1}{\scalebox{5.2}[1.15]{\textcolor{red}{$\bigcirc$}}}{O}{c}{F}{T}{L}%
}
\usepackage{cancel}

\usepackage{subcaption}
\usepackage{cleveref}
\algnewcommand\algorithmicinput{\textbf{Input:}}
\algnewcommand\algorithmicoutput{\textbf{Output:}}
\algnewcommand\Input{\item[\algorithmicinput]}%
\algnewcommand\Output{\item[\algorithmicoutput]}%
\algdef{SE}[SUBALG]{Indent}{EndIndent}{}{\algorithmicend\ }%
\algtext*{Indent}
\algtext*{EndIndent}
\usepackage{xcolor}
\makeatletter
\renewcommand\@makefntext[1]{%
    \setlength\parindent{1em}%
    \noindent
    \mbox{\textsuperscript{\@thefnmark}\,}{#1}}
\makeatother

\begin{document}
\author{Jiongli Zhu}
\affiliation{%
  \institution{University of California, San Diego}
}
\email{jiz143@ucsd.edu}

\author{Sainyam Galhotra}
\affiliation{%
  \institution{Cornell University}
}
\email{sg@cs.cornell.edu}

\author{Nazanin Sabri}
\affiliation{%
  \institution{University of California, San Diego}
}
\email{nsabri@ucsd.edu}

\author{Babak Salimi}
\affiliation{%
  \institution{University of California, San Diego}
}
\email{bsalimi@ucsd.edu}

\title{
Consistent Range Approximation for Fair Predictive Modeling}

\begin{abstract}
This paper proposes a novel framework for certifying the fairness of predictive models trained on biased data.  It draws from query answering for incomplete and inconsistent databases to formulate the problem of consistent range approximation (CRA) of fairness queries for a predictive model on a target population. The framework employs background knowledge of the data collection process and biased data, working with or without limited statistics about the target population, to compute a range of answers for fairness queries. Using CRA, the framework builds predictive models that are certifiably fair on the target population, regardless of the availability of external data during training. The framework's efficacy is demonstrated through evaluations on real data, showing substantial improvement over existing state-of-the-art methods.
\end{abstract}

\settopmatter{printfolios=true}

\maketitle

\section{Introduction}
\label{section:introduction}
\input{Section_Tex_Files/introduction}
\section{Preliminaries and Background}
\label{section:background}
We now review the background on ML, causality, and selection bias. Table \ref{table:our_notation} shows the notation we use.
\input{Section_Tex_Files/background}

\input{Section_Tex_Files/bounds}
\input{Section_Tex_Files/conditions}

\section{Experiments}

\input{Section_Tex_Files/experiments}
\section{Related Work}\label{section:relatedworks}

\input{Section_Tex_Files/related_work}


\input{Section_Tex_Files/discussion}


\bibliographystyle{ACM-Reference-Format}
\bibliography{main}

\end{document}

%% file: Section_Tex_Files/macros.tex
\usepackage{amsfonts}
\newcommand{\popul}{\Delta}
\newcommand{\auxdata}{\mc A_{\underlyingDist}}
\newcommand{\GAPair}{\mc I_{\underlyingDist}}
\usepackage[multiple]{footmisc}
\usepackage{booktabs}
\usepackage{dsfont}
\usepackage{xspace}
\usepackage{tikz}
\usetikzlibrary{shapes,arrows,calc}

\usepackage{tikz}
\usepackage[utf8]{inputenc}
\usepackage{balance}
\usepackage{siunitx}
\usepackage{multirow}
\usepackage{graphicx}
\usepackage{listings}
\usepackage{commath}
\usepackage{epstopdf}
\usepackage{hyperref}
\usepackage{color}
\usepackage{url}
\usepackage{algorithmicx}
\usepackage[noend]{algpseudocode}

\usepackage{caption}
\usepackage{alltt}
\usepackage{mathrsfs}
\usepackage{float}
\usepackage{subcaption}
\usepackage{cuted}
\usepackage{graphicx,fancyvrb}
\usepackage{amsmath}
\usepackage{booktabs}

\usepackage{mathtools}
\usepackage{caption}
\usepackage{float}
\usepackage{capt-of}
\usepackage{booktabs}
\usepackage{colortbl}
\usepackage{xcolor}
\usepackage{xfrac}
\usepackage{footnote}

 \usepackage{enumitem}
\usepackage{array}
\usepackage{arydshln}
\newif\ifextended
\newif\ifrevision

\usepackage{amsmath,amsfonts}

\definecolor{mygreen}{rgb}{0,0.6,0}
\definecolor{myred}{rgb}{0.6,0,0}
\definecolor{mygray}{rgb}{0.5,0.5,0.5}
\definecolor{mymauve}{rgb}{0.58,0,0.82}
\definecolor{myblue}{rgb}{0,0,1}
\definecolor{initblockcolor}{HTML}{F5D590}
\definecolor{decblockcolor}{HTML}{DBE3B5}
\definecolor{propblockcolor}{HTML}{BAD6EC}

\newcommand{\reva}[1]{{#1}}
\newcommand{\revb}[1]{{#1}}
 \newcommand{\revc}[1]{{#1}}
\newcommand{\revd}[1]{{#1}}

\usepackage{listings}
\lstnewenvironment{VerbatimText}[1][]{
    
    \lstset{fancyvrb=true,basicstyle=\footnotesize,captionpos=b,xleftmargin=2em,#1}
}{}

\usepackage{booktabs}
\usepackage{pgfplots}
\usepackage{pgfplotstable}

\PassOptionsToPackage{hyphens}{url}\usepackage{hyperref}

\newcommand{\cg}{ G}

\usepackage{amsmath}

\usepackage{bbm}

\newcommand{\Pa}{\boldsymbol{Pa}}
\newcommand{\pa}{\boldsymbol{Pa}}

\newcommand{\pr}{{\tt \mathrm{Pr}}}
\newcommand{\PrPos}{{\tt \mathrm{Pr}}^+}
\newcommand{\sys}{\textsc{Crab}\xspace}
\newcommand{\sysBound}{\textsc{Crab}-$\emptyset$\xspace}
\newcommand{\sysComplete}{\textsc{Crab}-\textsc{Suff}\xspace}
\newcommand{\sysMissA}{\textsc{Crab}-\textsc{NoA}\xspace}
\newcommand{\sysMissPartU}{\textsc{Crab}-\textsc{NoU}\xspace}

\newcommand{\CUBMissPartU}[1]{\textsc{CUB}-\textsc{NoU}-#1\xspace}

\newcommand{\lfr}{\textsc{Zemel}\xspace}
\newcommand{\ipw}{\textsc{Cortes}\xspace}
\newcommand{\minmax}{\textsc{Rezaei}\xspace}
\newcommand{\reweighing}{\textsc{Kamiran}\xspace}
\newcommand{\advdebias}{\textsc{Zhang}\xspace}
\newcommand{\covofdist}{\textsc{Zafar}\xspace}
\newcommand{\orig}{\textsc{Orig-SB}\xspace}
\newcommand{\origtest}{\textsc{Orig}-{NoSB}\xspace}

\newcommand{\bigCI}{\indep}

\newcommand{\nindep}{\mbox{$\not\!\perp\!\!\!\perp$}}
\newcommand{\indep}{\mbox{$\perp\!\!\!\perp$}}

\newcommand{\mc}[1]{\mathcal{#1}}

\newcommand{\ignore}[1]{}

\newcommand*{\rom}[1]{\expandafter\@slowromancap\romannumeral #1@}

\newcommand{\babak}[1]{{\texttt{\color{red} Babak: [{#1}]}}}

\newcommand{\Uvar}{\boldsymbol U}

\newcommand{\Yvar}{Y}
\newcommand{\Xvar}{\boldsymbol X}

\newcommand{\Allvar}{\boldsymbol X}
\newcommand{\dsep}{\mid}
\newcommand{\young}{Young}
\newcommand{\old}{Old}
\newcommand{\NPos}[1]{\texttt{N}_{#1}^{+}}

\newcommand{\seCondDemoParityDist}[1]{\hat{\digamma}(#1)}

\newcommand{\CondDemoParityDist}[2]{\digamma_{\classifier, {#1}}(#2)}

\newcommand{\sCondDemoParityDist}[1]{\digamma(#1)}

\newcommand{\CondDemoParity}[1]{\digamma_{\classifier, {#1}}(\underlyingDist)}
\newcommand{\CondDemoParityb}[1]{\digamma_{\classifier, {#1}}(\popul)}
\newcommand{\sCondDemoParity}{\digamma(\underlyingDist)}
\newcommand{\sCondDemoParityb}{\digamma(\popul)}

\newcommand{\fairapp}{(\classifier,  \digamma, \bdata, \GAPair)}

\usepackage{amsmath}

\newcommand{\dist}{{\underlyingDist}}

\newenvironment{proofsketch}{%
  \renewcommand{\proofname}{Proof Sketch}\proof}{\endproof}

\newcommand{\RNum}[1]{\uppercase\expandafter{\romannumeral #1\relax}}

\newcommand{\mb}[1]{{\mathbf{#1}}}
\newtheorem{defn}{Definition}[section]
\newtheorem{problem}{Problem}[section]

\newtheorem{col}[defn]{Colollary}

\newcommand{\proj}[1]{{\Pi}}
\newcommand{\sel}[1]{{\sigma}}

\newcommand{\cut}[1]{}
\newcommand{\eat}[1]{}

\usepackage[multiple]{footmisc}

\theoremstyle{remark}

\usepackage{amsmath}

\newcommand{\features}{\ensuremath{\boldsymbol{X}}}
\newcommand{\classifier}{\ensuremath{h}}
\newcommand{\bayesclassifier}{\ensuremath{h}^*}

\newcommand{\Dom}[1]{\textsc{Dom}(#1)}
\newcommand{\labelDom}{Y}

\newcommand{\underlyingDist}{\Omega}

\newcommand{\data}{D_{\dist}}

\newcommand{\lowerbound}{{\text{CLB}}}
\newcommand{\upperbound}{\text{CUB}}

\newcommand{\bdata}{D_{\popul}}
\newcommand{\dtrain}{\bdata^{tr}}
\newcommand{\dtest}{D_{\underlyingDist}^{ts}}

\newcommand{\admis}{\boldsymbol{a}}
\newcommand{\Admis}{\boldsymbol{A}}
\newcommand{\fairdef}{conditional statistical parity\xspace}

\newcommand{\datap}{\boldsymbol{x}}

\newcommand{\alabel}{y}
\newcommand{\selection}{C}
\newcommand{\cdag}{G}
\newcommand{\bcdag}{\mc G}

\newcommand{\repairs}{\ensuremath{\mathsf{Repairs}}\xspace}

\newcommand{\cra}{CRA}

\newcommand{\fairnessquery}{\digamma}

\newcommand{\protectedAttr}{s}
\newcommand{\protG}{s_0}
\newcommand{\privG}{s_1}
\newcommand{\ProtectedAttr}{S}
\newcommand{\error}{Loss}
\DeclareMathOperator{\EX}{\mathbb{E}}

\newcommand{\SelectionVar}{C}

\usepackage[linewidth=0.7pt]{mdframed}

%% file: Section_Tex_Files/introduction.tex

\revb{

Algorithmic decision-making systems are increasingly prevalent in critical domains, highlighting the importance of fairness. The objective is to ensure equal treatment across diverse groups based on sensitive attributes. Consequently, research in machine learning (ML), data management, and related fields has grown to address algorithmic fairness~\cite{balayn2021managing, islam2022through}. This paper particularly focuses on predictive modeling, aiming to train ML models that provide accurate predictions while ensuring fairness. Various metrics, such  equality of odds, are used to evaluate fairness~\cite{corbett2017algorithmic, pleiss2017fairness, hardt2016equality}.

Traditional fairness methods in predictive modeling can be categorized as in-processing or pre-/post-processing techniques~\cite{calders2010three,nabi2018fair,zhang2021omnifair,feldman2015certifying,pmlr-v65-woodworth17a,salimi2019interventional,galhotra2022causal}. However, these methods often assume that the training data is representative of the target population~\cite{islam2022through}. In practice, {\bf biases in data and quality issues} during data collection and preparation can distort the underlying data distribution, rendering it no longer representative of the target population. As a result, deploying these models in the target population may lead to unfair and inaccurate predictions~\citep{islam2022through,goel2021importance,mishler2022fair,barrainkua2022survey,kallus2018residual}.}

\reva{A significant issue in predictive models is \textbf{selection bias}, resulting from training data selection based on specific attributes, which creates unrepresentative datasets. This problem is prevalent in sensitive areas like predictive policing, healthcare, and finance, attributed to data collection costs, historical discrimination, and biases~\cite{kho2009written,greenacre2016importance,bethlehem2010selection,culotta2014reducing}. For example, in predictive policing, the data is biased as it is gathered exclusively from police interactions, which are influenced by the sociocultural traits of the officers~\cite{lange2005testing, gelman2007analysis}. Similarly, in healthcare, selection bias occurs when data is relied upon from individuals who are hospitalized or have tested positive, leading to disproportionate effects on racial, ethnic, and gender minorities due to barriers in healthcare access~\cite{access_to_health_care,cameron2010gender,zhao2020prediction,schwab2020clinical}.}

\vspace{-0.1cm}
\reva{\begin{example}
\label{ex:predictive-policing}
Consider the dataset in Table~\ref{tb:impossible}, which represents a sample collected from a population of individuals. The objective is to train a ML model to predict future crime risk based on various features while ensuring fairness according to the equality of odds principle. This principle aims to maintain similar true positive and false positive rates across different protected and sensitive groups.  However, our access to data is limited to that collected by police departments, resulting in a biased dataset. The dataset only includes records selected by law enforcement, indicated by the variable $C$, while records where $C=0$ are not available (highlighted in red). 

Using incomplete data for training an ML model has two significant ramifications. First, the selected subset of data may exhibit a spurious correlation between certain attributes (e.g., race, gender, age) and the label, which either does not exist or is not as significant in the complete dataset. This means that a classifier trained on this incomplete dataset learns this misleading correlation, resulting in subpar performance when applied to the entire dataset in terms of accuracy and fairness. For example, in the given case, although age and the training labels are not correlated in the complete data, they become correlated in the selected subset of data.

Second, solely relying on existing fair ML techniques to train a model that is fair on a selected subset of data does not guarantee fairness on the complete data or the target population where the model will be deployed. In the provided example, although the model's predictions on the selected subset demonstrate accuracy and fairness according to the equality of odds principle with respect to race, its performance deteriorates when applied to the complete dataset, resulting in inaccurate and unfair predictions (see Example~\ref{ex:impossibility} for more details). 

\end{example}}
%



\ignore{
\vspace{-0.1cm}
\begin{example}
\label{ex:predictive-policing}
Obtaining unbiased data from other sensitive domains such as healthcare and finance is a major challenge. In healthcare, studies are often conducted on data collected from individuals who have already been admitted to hospitals or have a positive test result, which can lead to selection bias as certain groups may have less access to healthcare~\cite{access_to_health_care,cameron2010gender}. This can be seen in the context of COVID-19, where studies are conducted on data collected from individuals who have a positive PCR test or have been admitted to hospitals~\cite{zhao2020prediction,schwab2020clinical}. This leads to bias as racial and ethnic minorities~\cite{access_to_health_care} and gender minorities~\cite{cameron2010gender} have been shown to have less access to healthcare.
In finance, data collected from financial institutions may also be subject to selection bias, such as those used for credit risk assessment or loan approval. For example, individuals from certain racial or socio-economic backgrounds may be less likely to have access to financial services or to have a credit history, leading to bias in financial datasets~\cite{beck2009access,banasik2003sample}. 
\end{example}
}

\reva{
Recent efforts to mitigate selection bias in ML often rely on accessing unbiased samples from the target population (e.g., \cite{sugiyama2007covariate,cortes2008sample,bareinboim2012controlling,liu2014robust,huang2006correcting,rezaei2020robust}). However, obtaining unbiased samples from sensitive domains like predictive policing, healthcare, and finance poses significant challenges due to inherent biases specific to each domain. Furthermore, methods that rely on biased samples have demonstrated poor real-world performance \cite{cortes2008sample,liu2014robust}, often neglecting fairness concerns in predictive modeling.



The common theme in data collection across sensitive domains like predictive policing, healthcare, and finance is that while obtaining unbiased data may be infeasible, it's feasible to acquire background knowledge about the {\bf data collection process}. In predictive policing, data factors include demographic characteristics, sociocultural traits, and residence. In healthcare, datasets reflect age, race, socio-economic status, and the selection process. For finance, credit risk or loan approval datasets are influenced by access to financial services, income, employment status, and location. Additionally, it's possible to gather partial information about the target population from \textbf{external data sources} such as census data, open knowledge graphs, and data lakes. In healthcare, external sources like government databases provide unbiased aggregated demographic, socioeconomic, and geographic information, including healthcare resources access and insurance coverage. For predictive policing and finance, external sources like crime statistics, open data lakes, and credit bureau data offer unbiased insights into race, socio-economic status, and credit history. In finance, credit bureau data, financial institution databases, and government statistics provide valuable unbiased demographic, income, employment status, and location information.
}

{\scriptsize
\begin{table}[]
\centering
\resizebox{\columnwidth}{!}{%
\begin{tabular}{lccccll}
\toprule
{\textbf{Crime Type}} &
  { \textbf{Age}} &
  { \textbf{Race}} &
  { \textbf{ZIPCode}} &
  {\textbf{$\SelectionVar$}} &
  {\textbf{Y (label)}} &
  {\textbf{Prediction}} \\ \midrule
\rowcolor[HTML]{F7D0D0}
Arson & \young & Black & 90043 & 1 & High Risk & High Risk \\
\rowcolor[HTML]{F7D0D0}
Homicide & \young & Black & 90043 & 1 & High Risk & High Risk \\
\rowcolor[HTML]{F7D0D0}
Theft & \old & Black & 90043 & 1 & Low Risk & Low Risk \\
\rowcolor[HTML]{F7D0D0}
Robbery & \young & Black & 90043 & 1 & High Risk & High Risk \\
Car Break-in & \old & White & 90026 & 0 & High Risk & \Circle{Low Risk} \\
\rowcolor[HTML]{F7D0D0}
Assault & \old & White & 90026 & 1 & Low Risk & Low Risk \\
Theft & \young & White & 90026 & 0 & Low Risk & \Circle{High Risk} \\
Assault & \young & Black & 90026 & 0 & Low Risk & \Circle{High Risk} \\
\rowcolor[HTML]{F7D0D0}
Armed Robbery & \old & White & 90026 & 1 & High Risk & High Risk \\
\end{tabular}
}
\caption{\reva{\textmd{A toy dataset demonstrating selection bias in Example~\ref{ex:predictive-policing}. Circled cells correspond to wrong predictions.}}}
\label{tb:impossible}
\vspace{-1cm}
\end{table}
}

\reva{
In this work, we propose a novel framework called \textbf{C}onsistent \textbf{R}ange \textbf{A}pproximation from \textbf{B}iased Data (\sys) to address the challenge of constructing certifiably fair predictive models in the presence of selection bias, even when obtaining unbiased samples from the target population is not feasible. The key idea of \sys is to leverage background knowledge on the data collection process, encoded through a causal diagram representing the dependence between the selection of data points and their corresponding features. Moreover, \sys can incorporate external data sources that provide additional information about the target population to enhance its results. By understanding the data collection process, \sys formulates conditions that enable the training of predictive models that are certifiably fair on the target population. Unlike previous techniques that rely on unbiased samples and do not explicitly address fairness, \sys ensures fairness even in the absence of unbiased data during model training and testing.
}

\reva{
\begin{example} Continuing with example~\ref{ex:predictive-policing}, our system $\sys$ takes as input the incomplete data (the selected subset) in Table~\ref{tb:impossible}, along with background knowledge that indicates the selection of data points is dependent on individual Zipcodes and other socio-cultural traits. This information is encapsulated using a simple causal model that encodes that the variable $C$ is a function of factors such as Zipcode and other socio-cultural traits (This will be further elucidated in Section~\ref{sec:seelctionbias-back}, Figure~\ref{fig:police-dag}). Our system can also incorporate potential external data sources that contain unbiased information about the distribution of sensitive attributes in Zipcodes and so forth. Subsequently, the system trains an ML model using the incomplete data, ensuring that it will be fair on the complete data and, consequently, the target population.
\end{example}
}



\reva{
Selection bias presents a fundamental challenge in training fair predictive models, as it prevents the accurate measurement of {\bf fairness queries} - aggregate queries used to quantify fairness violations in an ML model on the target population, such as equality of odds. For instance, in the case of equality of opportunity, the fairness query is an aggregate query in the range [0,1] that assesses the disparity in the likelihood of a positive outcome between privileged and protected groups. This problem arises because {\bf selection bias results in incomplete and inconsistent data}, making fair ML modeling essentially a {\bf data management problem} concerning query answering from incomplete data. While query answering from inconsistent and incomplete data has been extensively studied in the database field~\cite{bertossi2006consistent, fan2012foundations,dixit2021consistent}, it hasn't been specifically addressed for biased data, apart from ~\cite{orr2020sample}, which assumes access to unbiased data samples.

To tackle this issue, we draw insights from data management to introduce and formalize the problem of {\bf Consistent Range Approximation (\cra) of fairness queries from biased data}. This approach aims to approximate the fairness of an ML model on a target population using background knowledge about the data collection process and limited or no  information from external data sources. Inspired by Consistent Query Answering in databases~\cite{bertossi2006consistent,dixit2021consistent}, $\cra$ considers the space of all possible repairs that are consistent with the available information. It uses this to compute a range for fairness queries such that the true answers are guaranteed to lie within the range. We refer to this as the \textit{consistent range}. {\bf (Section~\ref{sec:cra})}}

We present a closed-form solution for the problem of Consistent Range Answer (\cra) fairness queries in predictive modeling. Our analysis focuses on a class of aggregate queries that capture different notions of algorithmic fairness, such as statistical parity, equality of odds, and conditional statistical parity. \reva{We demonstrate that the consistent range can be efficiently calculated by incorporating varying levels of information about the target population from external data sources.}  This approach allows us to estimate the fairness of a model on the target population using biased data. Our results facilitate both the verification of approximate fairness and the training of certifiably fair models on all populations consistent with the available information about the target population, including the target population itself. The ability of \cra\ to accommodate varying levels of external data sources makes it a practical solution for addressing selection bias ({\bf Section~\ref{section:bounds}}).


Furthermore, we conduct a theoretical analysis of the impact of selection bias on the fairness of predictive models and establish necessary and sufficient conditions on the data collection process under which selection bias leads to unfair predictive models. Our results indicate that selection bias does not necessarily lead to unfair models, and in situations where it does, existing techniques are often inapplicable for training fair classifiers. This highlights the importance of addressing selection bias in the data management stage, rather than relying on post-processing methods {\bf (Section~\ref{section:introducing_bias_out_of_the_blue})}.


We evaluate $\sys$ on both synthetic and real data. Our findings show that when selection bias is present: (1) existing methods for training predictive models result in unfair models. (2) In contrast, \sys\ develops predictive models that are guaranteed to be fair on the target population. (3) Even when limited external data about the target population is available, \sys\ still produces fair and highly accurate predictive models. (4) In certain situations, enforcing fairness can also improve the performance of predictive models. (5) Interestingly, the predictive models developed by \sys\ in the presence of limited external data outperform those trained using current methods that have access to complete information about the target population~{\bf (Section~\ref{sec:experiments})}.

This paper is organized as follows: In Section~\ref{section:background}, we provide background information on fairness, causality, and selection bias. In Section~\ref{sec:cra}, we introduce and study the problem of \cra\ fairness queries, and we establish conditions under which it is possible to train certifiably fair ML models with varying access to external data sources about the target population. In Section~\ref{section:introducing_bias_out_of_the_blue}, we establish sufficient and necessary conditions under which fairness leads to unfair predictive models. Finally, in Section~\ref{sec:experiments}, we provide experimental evidence that \sys outperforms the state-of-the-art methods for training fair ML models and learning from biased data.

%% file: Section_Tex_Files/background.tex
\begin{table}[h] \scriptsize
\centering
 \begin{tabular}{||c | c||} 
 \hline
 Symbol & Meaning\\ [0.5ex] 
 \hline\hline
 $X,Y,Z$ & Attributes (features, variables) \\ 
 $\boldsymbol{X}, \boldsymbol{Y}, \boldsymbol{Z}$ & Sets of attributes \\
 $x, y, z$ & Attribute values\\
 $\Dom{X}$ & Domain of an attribute\\ 
 $\cdag$ & A causal diagram\\
 $D^{tr}, D^{ts}$ & Training/testing datasets\\
 $\data$ & Data sampled from distribution $\dist$\\ 
 $MB(X)$ & Markov Boundary of a variable $X$\\ 
 $\classifier(\boldsymbol{x})$ & A classifier\\[1ex]
 \hline
 \end{tabular}
 \caption{Notation used in this paper.}
 \label{table:our_notation}
 \vspace{-.8cm}
\end{table}


In this work, we focus on the problem of {\em binary classification}, which has been the primary focus of the literature on algorithmic fairness. Consider a population or data distribution $\underlyingDist$ with support $\features \times \labelDom$, where $\features$ denotes a set of discrete and continuous features and $\Dom{\labelDom} = \{0, 1\}$ represents some binary outcome of interest (aka the \textit{target attribute}). A classifier $\classifier: \Dom{\features} \to \{0, 1\}$ is a function that predicts the unknown label $y$ as a function of observable features $\boldsymbol{x}$. The quality of a classifier $\classifier$ can be measured using the {\em expected loss}, also known as the {\em risk}, i.e., $\error(h)=\EX_{\underlyingDist}[L(\classifier(\datap), y)]$, where $L(\classifier(\datap), y)$ is a {\em loss function}
that measures the cost of predicting $\classifier(\datap)$ when the true value is $y$. In this paper, we focus on the {\em zero-one loss}, i.e., $L(\classifier(\datap), y)=\mathbbm{1}(\classifier(\datap)\neq\alabel)$. 
A learning algorithm aims to find a classifier $\bayesclassifier \in \mc{H}$ that has a minimum loss, i.e., for all classifiers $\classifier \in \mc{H}$, it holds that $\error(\bayesclassifier) \leq \error(\classifier)$, where $\mc{H}$ denotes the hypothesis space. In the case of the zero-one loss, the optimal classifier $\bayesclassifier$ is called the {\em Bayes optimal classifier} and is given by $\bayesclassifier(\datap)$ iff $\Pr(y=1\mid \datap) \geq \frac{1}{2}$.
Since $\underlyingDist$ is unknown, we cannot calculate $\error(.)$ directly. Instead, given an i.i.d. sample $D^{tr}_\underlyingDist=\{(\boldsymbol{x}_i, y_i)\}_{i=1}^{n}$ from $\underlyingDist$, the {\em empirical loss}  $ \frac{1}{n}\sum_{i=1}^{n} L(\classifier(\datap), y)$ is typically used to estimate the expected loss. However, minimizing the empirical loss for the zero-one loss function is NP-hard due to its non-convexity. As a result, a convex surrogate loss function is used by learning algorithms. 
A surrogate loss function is said to be {\em Bayes-risk consistent} if its corresponding empirical minimizer converges to the Bayes optimal classifier when training data is sufficiently large~\citep{bartlett2006convexity}. 




\par {\bf Algorithmic fairness.}
Algorithmic fairness, particularly in the context of a binary classifier $\classifier$ with a protected attribute $\ProtectedAttr \in \features$ (e.g., gender or race), aims to ensure non-discriminatory predictions. In this setting, we define $\classifier(.)=1$ as a favorable prediction and $\classifier(.)=0$ as an unfavorable one. For simplicity, we denote $\Pr(\classifier(.)=1)$ as $\PrPos(\classifier(.))$. We assume that $\Dom{\ProtectedAttr} = \{\protG, \privG\}$, representing a privileged group ($\privG$) and a protected group ($\protG$). We focus on {\em Conditional Statistical Parity}\cite{zafar2017fairness}, which requires equal positive classification probabilities for both groups, conditioned on admissible features $\Admis$. These admissible features, such as a person's past convictions in predictive policing, are non-discriminatory and thus permissible for decision-making. Formally, this definition necessitates that for all admissible feature values $\admis \in \Dom{\Admis}$, we have $\PrPos_{\underlyingDist }(\classifier(\boldsymbol{x})\mid \protG, \Admis=\admis)=\PrPos_{\underlyingDist }(\classifier(\boldsymbol{x})\mid\privG, \Admis=\admis)$.


Conditional statistical parity is a versatile fairness notion that captures many other fairness principles. When the set of admissible features is empty ($\Admis=\emptyset$), it reduces to {\em Statistical Parity}~\cite{dwork2012fairness}, which requires equal positive classification probabilities for the protected and privileged groups. Conversely, when the set of admissible features includes all available features ($\Admis=Y$), it encompasses Equality of Opportunity~\cite{hardt2016equality}, which demands equal positive prediction probabilities for both groups among individuals who should have received a favorable outcome based on the ground truth. Therefore, satisfying conditional statistical parity addresses both statistical parity and equality of opportunity. Furthermore, the methods developed in this paper can be extended and adapted to handle other fairness notions, such as causal notions~\citep{salimi2019capuchin}.

\vspace{-0.3cm}
\subsection{Background on Selection Bias}
\label{sec:seelctionbias-back}

Selection bias occurs when the selection of a data point in a sample from a data distribution is not based on randomization, but rather on certain attributes of the data point. This results in a sample that is not representative of the data distribution. To address this issue, we use causal diagrams to encode background knowledge about the data collection process and its potential biases. This approach allows us to model and analyze selection bias in a principled way. Next, we give a short overview of how we employ causal diagrams.

\vspace{-0.3cm}
\subsubsection{Causal diagrams}  A {\em causal diagram} is a directed graph that represents the causal relationships between a collection of {\em observed} or {\em unobserved} (latent) variables $\Allvar$ and models the underlying process that generated the observed data. Each node in a causal diagram corresponds to a variable $X\in \Allvar$, and an edge between two nodes indicates a potential causal relationship between the two variables.
The set of all parents of a variable $X$ is denoted by $\Pa(X)$.

\noindent \textbf{$d$-separation and collider bias.} {\em Causal diagrams encode a set of conditional independences } that can be read off the graph using  \textit{$d$-separation}~\cite{pearl2009causality}. A {\em path} is a sequence of adjacent arcs, e.g., $(Race \rightarrow \boldsymbol{Z} \leftarrow ZIPCode)$ in Figure~\ref{fig:police-dag:a}. Two nodes are {\em $d$-separated} by a set of variables $\boldsymbol{V}_m$ in causal diagram $G$, denoted $(V_l \indep V_r \mid_d \boldsymbol{V}_m)$ if for every path between them, {\em one} of the following conditions holds: \reva{(1) the path contains a {\em chain} ($V_l \rightarrow V \rightarrow V_r)$  or a {\em fork} ($V_l \leftarrow V \rightarrow V_r)$ such that $V \in \boldsymbol{V}_m$, and (2) the path contains a {\em collider} ($V_l \rightarrow V \leftarrow V_r)$ such that $V  \not\in \boldsymbol{V}_m$,  and no descendants of $V$ are on $\boldsymbol{V}_m$.} For example, $Y$ and $R$ are {\em $d$-separated} by $N$ and $\boldsymbol{X}$ in $\cg$. Key to $d$-separation is that conditioning on a collider (common effect) can induce a spurious correlation between its parents (causes), a phenomenon known as {\em collider bias}~\cite{pearl2009causality}.


\noindent \textbf{Conditional Independence.} 
A data distribution $\underlyingDist$ is said to be {\em Markov compatible}, or simply compatible, with a causal diagram $\cdag$ if d-separation over the $\cdag$ implies conditional independence with respect to  $\underlyingDist$. More formally, ($\boldsymbol{X} \bigCI \boldsymbol{Y} \mid_d  \boldsymbol{Z})  \implies (\boldsymbol{X} \bigCI_\underlyingDist\boldsymbol{Y} \mid  \boldsymbol{Z} $),
where $(\boldsymbol{X} \bigCI_\underlyingDist\boldsymbol{Y} \mid  \boldsymbol{Z})$ means $\boldsymbol{X}$ is independent of $\boldsymbol{Y}$ conditioned on $\boldsymbol{Z}$ in $\underlyingDist$. If the converse also holds, i.e., ($\boldsymbol{X}\bigCI_{\underlyingDist} \boldsymbol{Y} \mid\boldsymbol{Z}) \implies (\boldsymbol{X}\bigCI \boldsymbol{Y} \mid_{d}\boldsymbol{Z}$), $\underlyingDist$ is considered {\em faithful} to $\cdag$. 
%
\begin{example}
\label{ex:diagram}
Figure \ref{fig:police-dag}a shows a simplified causal model for Example~\ref{ex:predictive-policing}) with variables $Y$: drug use; $\boldsymbol{W}$: variables such as income, education, job that are deemed to causally affect drug use; $\boldsymbol{Z}$: sociocultural traits, zip code, and race. 
For a distribution $\underlyingDist$ compatible with this causal diagram, it holds that $(Race \indep_{\underlyingDist} Y)$ because $Race$ and $Y$ are $d$-separated by an empty set since the path ($Race \rightarrow \boldsymbol{Z} \leftarrow ZIPCode \rightarrow  \boldsymbol{W} \rightarrow Y$) is closed at a collider node $\boldsymbol{Z}$.  
\end{example} 
\vspace{-0.3cm}
\subsubsection{Data collection Diagrams}  We use causal diagrams to represent selection bias in data collection. Given a {\bf target population} $\underlyingDist$ that is faithful to a causal diagram $\cdag$, we can model a biased data collection process (where the selection of data points depends on a set of variables $\boldsymbol{V} \subseteq \Allvar\cup\{Y\}$) using a data collection diagram $\bcdag$. This is achieved by augmenting $\cdag$ with a selection node $\selection$, where $\boldsymbol{V}$ constitutes the parents of $\selection$, i.e. $\boldsymbol{V}=\pa(\selection)$. In this way, the collected data $\bdata$ can be seen as an i.i.d sample from a {\em biased data distribution} $\popul$ that is compatible with $\bcdag$, such that $\pr_{\popul}(\boldsymbol{x},y)= \pr_{\underlyingDist}(\boldsymbol{x},y \mid \SelectionVar=1)$. Additionally, conditioning on $\selection$ in the biased data distribution may result in a spurious correlation between variables for which $\SelectionVar$ is a collider in a path between them. We provide an example to illustrate this concept.
\begin{figure}[h]
\vspace{-.3cm}
\centering
  \begin{subfigure}{0.3\linewidth}
    \includegraphics[width=\linewidth]{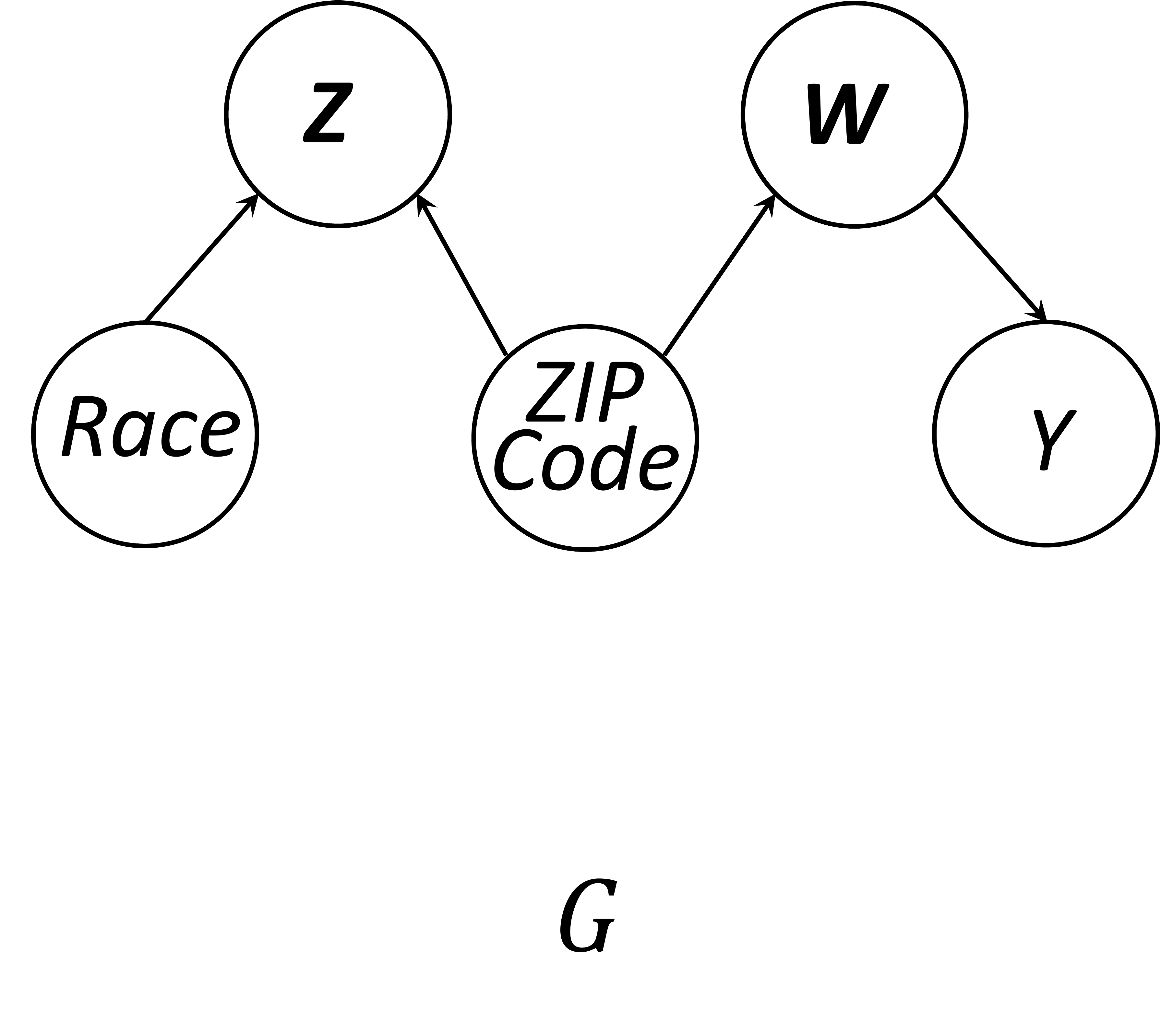}
    \vspace{-.55cm}
    \caption{}
    \label{fig:police-dag:a}
  \end{subfigure}
  \hspace{0.6cm}
  \begin{subfigure}{0.3\linewidth}
    \includegraphics[width=\linewidth]{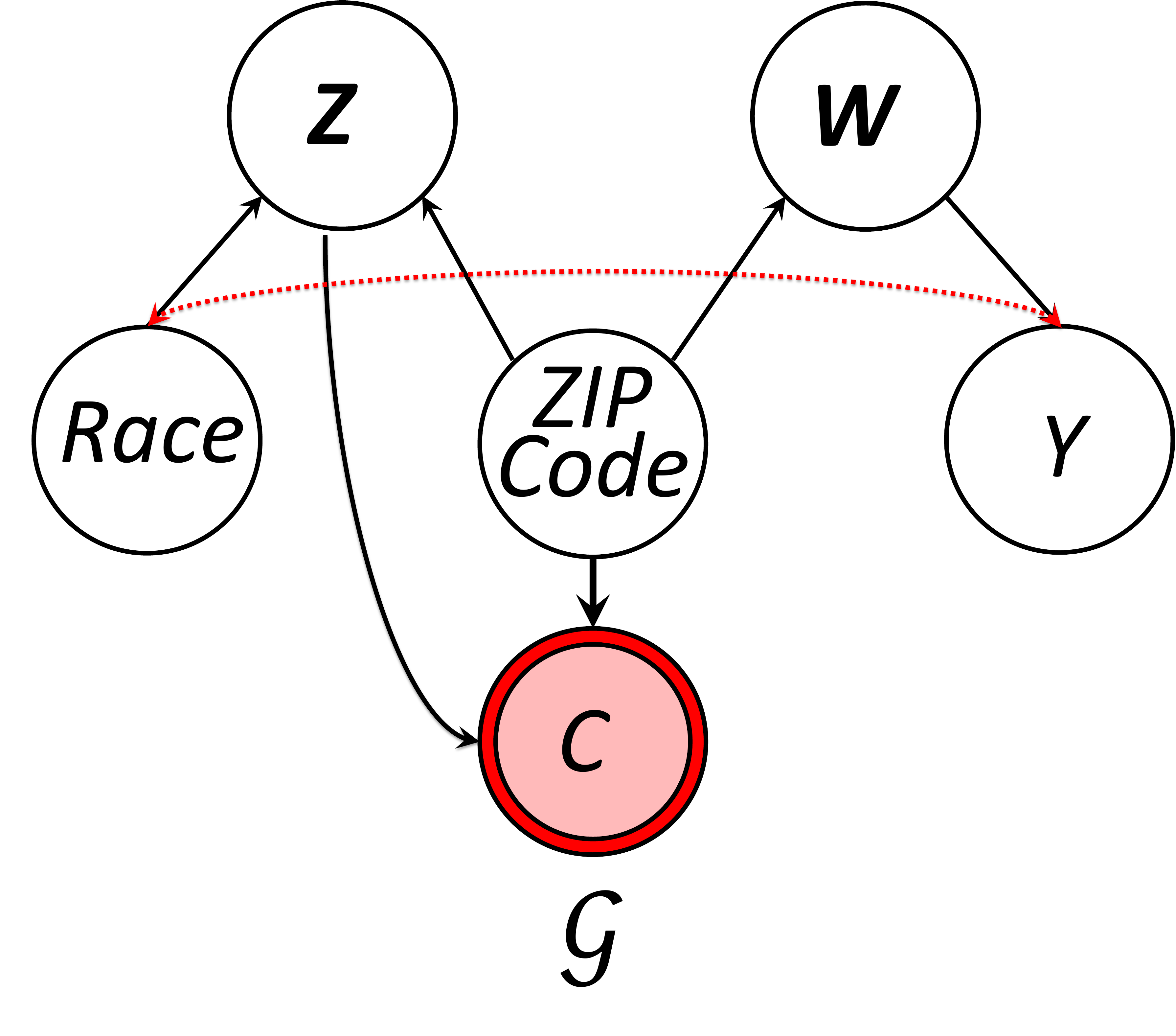}
    \vspace{-.55cm}
    \caption{}
    \label{fig:police-dag:b}
  \end{subfigure}
  
\vspace{-0.4cm}
\caption{\reva{\textmd{(a) A causal diagram $\cdag$ for predictive policing in Examples~\ref{ex:diagram}. (b) Data collection diagram $\bcdag$ of police data in Example~\ref{ex:predictive-policing2} in the presence of selection bias.}}}
\label{fig:police-dag}
\end{figure}
\vspace{-0.6cm}
\begin{example} (Example~\ref{ex:diagram} continued) 
\label{ex:predictive-policing2} Figure \ref{fig:police-dag}b shows a data collection diagram for a scenario in which the selection of data points into a sample depends on the neighborhood that is more regularly patrolled and individuals' sociocultural traits, as indicated by arrows from the $\boldsymbol{Z}$ and $ZIPCode$ variables to the selection variable $\SelectionVar$. A sample of data collected according to this biased data collection process can be seen as a sample from a biased data distribution $\pr_{\popul}(\Allvar)= \pr_{\underlyingDist}(\Allvar \mid \SelectionVar=1)$, where $\Allvar=\{\boldsymbol{W},\boldsymbol{Z},Y, Race, $  $ ZIPCode\}$. Furthermore, since $\SelectionVar$ is a collider in the path ($Race \rightarrow \boldsymbol{Z} \rightarrow \SelectionVar \leftarrow ZIPCode \rightarrow \boldsymbol{W} \rightarrow Y$) between $Race$ and $Y$, conditioning on the selection variable $\SelectionVar$ induces a spurious correlation between $Race$ and $Y$ that does not exist in the target distribution $\underlyingDist$, but exists in the biased data $\popul$, due to the collider bias, as indicated by the red bidirectional arrow between $Race$ and $Y$ in the diagram.
Specifically, while in the target data distribution $\underlyingDist$ it holds that ($Race \indep_{\underlyingDist} Y$), in the biased data distribution $\popul$ ($Race \nindep_{\popul} Y$).
Hence, a sample $\bdata$ from $\popul$ may exhibit spurious correlations between $Race$ and $Y$ that may lead unfair predictive models trained on $\bdata$. 
Additionally,  the data collection diagram encodes the independence of race, $\boldsymbol{W}$, and $Y$ from the selection variable given $\boldsymbol{Z}$ and ZIP code. This means that the conditional distribution $\pr_{\underlyingDist}(Race, \boldsymbol{W},Y \mid \boldsymbol{Z},ZIPCode)$ of the target population can be calculated from biased data, and therefore biased data can provide some information about the target population's statistics. This is because $\pr_{\underlyingDist}(Race, \boldsymbol{W},Y \mid \boldsymbol{Z},ZIPCode)=\pr_{\underlyingDist}(Race, \boldsymbol{W},Y \mid \boldsymbol{Z},ZIPCode, \SelectionVar=1)=\pr_{\popul}(Race, \boldsymbol{W},Y \mid \boldsymbol{Z},ZIPCode)$. 
\end{example}

%% file: Section_Tex_Files/bounds.tex
\section{Fair Classification Under Selection Bias Using \sys}
\label{section:bounds}

The presence of selection bias can result in a discrepancy between the target distribution $\underlyingDist$ and the biased data distribution $\Delta$. This variance in the two distributions can result in a difference in the unfairness of a classifier evaluated on each. For instance, a classifier that is fair on the biased training data can still exhibit unfairness when deployed on the actual data distribution~\cite{mishler2022fair}.

\begin{example}\label{ex:impossibility}
Consider the classifier, $\classifier$(Age, Crime Type) = High Risk if (Crime Type = Armed Robbery) or (Age = \young), otherwise Low Risk, as presented in Example~\ref{ex:predictive-policing} using Table~\ref{tb:impossible}. This classifier, derived from biased data, seems fair in terms of equality of odds and appears accurate when applied to the selected subset used for training. Although it's feasible to devise a perfect and fair classifier based solely on crime type, this model capitalizes on the observed correlation between age and high risk within the selected subset, creating a seemingly flawless and fair model. However, when this classifier is applied to the entire dataset, it manifests as unfair, despite its predictions not being directly based on race.
\end{example}

Our goal is to develop ML models that can effectively handle selection bias while ensuring fairness during deployment on the target  distribution. We formally define this objective as follows:
\begin{defn}[Fair classifier]\label{def:fair_clf} A classifier $\classifier$ trained on a training data $\dtrain$ sampled under selection bias from a biased distribution $\popul$  to predict the class label $\Yvar$ is fair if it satisfies \fairdef\ on an (unseen) test data $\dtest$ that is a representative sample of the target distribution ${\underlyingDist}$. \end{defn}
\revb{
It is widely recognized in the literature that learning a fair classifier without any auxiliary information about the target distribution or the data collection process is practically impossible \cite{singh2021fairness,chen2022fairness,giguere2022fairness,wu2021learning,david2010impossibility}. Without any auxiliary information to serve as constraints, the target distribution may deviate significantly from the distribution observed in the training data. This is illustrated in Example~\ref{ex:impossibility}. The importance of having knowledge about the target distribution has been acknowledged in the literature \cite{schrouff2022maintaining,mukherjee2022domain,subbaswamy2019preventing}.}


To address this challenge, first we introduce the problem of consistent range approximation (\cra), which aims to approximate and bound the answers to fairness queries on a target population using biased data (Section~\ref{sec:cra}). Then, we study the problem of \cra\ in settings for which limited or no external data source about the target population is available (Section~\ref{section:recoverability:withnoExternalData}-\ref{section:recoverability:withExternalData}). 
\revb{The \cra\ framework for fairness queries can be combined with standard approaches, such as incorporating fairness constraints or regularization into the learning process. By adding upper bounds of fairness queries as constraints or regularization terms, the model is penalized to ensure it has an approximately zero consistent upper bound. This ensures fairness across all possible repairs and ultimately extends to the target population (Section \ref{sec:recoverability:regularizer})}.


\vspace{-0.2cm}
\subsection{
Consistent Range Approximation (\cra) of a Fairness Query}\label{sec:cra}




In this work, we utilize the total variation distance as a measure for evaluating the degree of fairness violation of a classifier $\classifier$, wrt. the fairness definitions outlined in Section~\ref{section:background}.\ifextended~\footnote{The total variation distance between two distributions $P$ and $Q$ is defined as $\text{TV}(P,Q) = \sup_{\mb x \in \Dom{\mb X}} \left| P(\mb x) - Q(\mb x) \right|$. It is known that the total variation distance is half of the $L_1$ norm between the distributions, i.e. $\text{TV}(P,Q) = \frac{1}{2} \sum_{\mb x \in \Dom{\mb X}} \left| P(\mb x) - Q(\mb x) \right|$~\cite{levin2017markov}}.\fi This approach encompasses a variety of other methods found in the literature for identifying discrimination (see, e.g., ~\cite{zhang2018equality,zhang-aaai18,ashurst2022fair, rezaei2020robust, zhang2021omnifair,zhang2018mitigating, moldovan2022algorithmic,bellamy2019ai,kozodoi2022fairness,zhang2021ignorance,stevens2020explainability,madras2018learning}). Specifically, we formulate the fairness query as follows:

%
\begin{defn}[Fairness Query]
We measure the fairness violation of a classifier $\classifier$ with respect to a given set of admissible attributes $\Admis$ and on a population $\dist$ with support $\features \times \labelDom$ using a {\em fairness query} defined as follows:
{ \small
\begin{flalign}
\indent&\begin{aligned}
&\CondDemoParityDist{\Admis}{\dist} \\
&= \frac{1}{ 2 \ \lvert\Admis\rvert}  \sum_{ \substack{y\in\Dom{\Yvar},\\ \admis\in\Dom{\Admis}} }\big\lvert\pr_{\dist }(\classifier(\boldsymbol{x})=y\mid \privG, \admis)-\pr_{\dist }(\classifier(\boldsymbol{x})=y\mid\protG, \admis)\big\rvert. \vspace{-1cm}
\end{aligned}&&
\label{eq:fairnessquery}
\end{flalign}
}
\vspace{-5mm}
\end{defn}

The fairness query in Eq~\eqref{eq:fairnessquery} quantifies the level of unfairness of a classifier $\classifier$ by comparing the average total variation distance between the conditional probability distributions $\PrPos_{\dist }(\classifier(\boldsymbol{x})\mid \privG, \admis)$ and $\PrPos_{\dist }(\classifier(\boldsymbol{x})\mid \protG, \admis)$ for all $\mb a \in \Dom{\Admis}$. A classifier $\classifier$ is considered $\epsilon-$fair on population $\dist$ if $\CondDemoParityDist{\Admis}{\dist} \leq \epsilon$. Notably, Eq~\eqref{eq:fairnessquery} measures discrimination for a $\boldsymbol x \in \Dom{\boldsymbol X}$. The overall unfairness can be calculated by taking the expectation over $\boldsymbol x$. It's important to note that a  classifier that satisfies conditional statistical parity on $\dist$ would have a value of $\CondDemoParityDist{\Admis}{\dist} = 0$. For brevity, in subsequent sections we will refer to the fairness query by omitting explicit references to $\dist$, $\classifier$, and $\Admis$ when the context is clear, e.g., using the symbol $\digamma$.

Our focus is on binary classification and, for simplicity, we assume $\PrPos_{\dist}(\classifier(\boldsymbol{x})\mid \privG, \admis)\geq\PrPos_{\dist}(\classifier(\boldsymbol{x})\mid\protG, \admis)$ holds for arbitrary $\admis\in\Admis$ throughout this section. However, this assumption is not restrictive and can be easily adapted during implementation without compromising the generality of our results. Under this assumption, the fairness query in Eq~\eqref{eq:fairnessquery} simplifies as follows:
{\small
\begin{equation}
\begin{aligned}
\sCondDemoParityDist{\dist}= \frac{1}{\lvert\Admis\rvert} \sum_{\admis\in\Dom{\Admis}} \PrPos_{\dist}(\classifier(\boldsymbol{x})\mid \privG, \admis) - \PrPos_{ \dist }(\classifier(\boldsymbol{x})\mid \protG, \admis)
\label{eq:binaryfairnessquery}
\end{aligned}
\end{equation}
}

In practice, the fairness query in Eq~\eqref{eq:fairnessquery} must be computed using data through an {\em empirical fairness query} denoted $\hat{\digamma}$, which is typically defined based on the empirical estimate as follows:

\begin{defn}[Empirical fairness Query]\label{def:empfairquery}
Fairness violation of a classifier $\classifier$ on a dataset $\data$ sampled from a distribution $\dist$ can be obtained by the following {\em empirical fairness query}:  
{\small
\begin{align}
&\seCondDemoParityDist{\data}= \frac{1}{\ \lvert\Admis\rvert} \sum_{\admis\in\Dom{\Admis}} \Bigg\lvert \frac{\sum_{\boldsymbol{x}\in\NPos{\privG, \admis}}{\classifier(\boldsymbol{x})} }{\lvert\NPos{\privG, \admis}\rvert}-
\frac{\sum_{\boldsymbol{x}\in\NPos{\protG, \admis}}{\classifier(\boldsymbol{x})} }{\lvert\NPos{\protG, \admis}\rvert}\Bigg\rvert
\label{eq:empiricalfairnessquery}
\end{align}
}
where $\privG$ and $\protG$ are the protected attributes, and $\NPos{s, \admis}$ denotes the set of data points in $\data$ with positive labels, protected attribute value $\ProtectedAttr=\protectedAttr$ and admissible attributes value $\Admis=\admis$. 
\end{defn}

\revb{Note that $\seCondDemoParityDist{\data}$ is an unbiased estimate of $\sCondDemoParityDist{\dist}$, and is often used as it can be estimated from the data (see, e.g., \cite{bechavod2017learning,konstantinov2021fairness,kamiran2012data,zhang2021omnifair} for details). In order to avoid sampling variability, in the subsequent, we assume samples are sufficiently large such that $\seCondDemoParityDist{D_\dist}\approx\sCondDemoParityDist{\dist}$ and use them interchangeably.

The problem lies in the fact that we do not have access to the true data, $\data$, and the only data we have is $\bdata$, which is drawn from a biased distribution $\Delta$. This means that the empirical fairness estimate, $\seCondDemoParityDist{\bdata}$, can provide a biased and incorrect estimate of the fairness of a classifier on the target population, $\sCondDemoParityDist{\underlyingDist}$. This is a concern even if the sample $\bdata$ is large, as $\sCondDemoParityDist{\popul}$ and $\sCondDemoParityDist{\underlyingDist}$ are generally not equal. Our aim is to approximate and bound $\sCondDemoParityDist{\underlyingDist}$ using the biased $\bdata$ and auxiliary information regarding the data collection process and target population.}


\begin{defn}[Auxiliary Information]
The {\em auxiliary information} $\GAPair=(\bcdag,\auxdata)$ is a tuple where $\bcdag$ is the data collection diagram that represents the underlying biased data collection process, and $\auxdata$ is a set of external data sources that can potentially provide {\em partial} information about the target population $\underlyingDist$.
\label{def:auxinfo}
\end{defn}




We now introduce our problem setup.
\begin{defn}[Fairness Query Application] A {\em fairness}\linebreak {\em query application} is a tuple $\fairapp$, where $\classifier$ is a classifier trained on a dataset $\bdata$ that was collected under selection bias and corresponds to a biased distribution $\popul$, $\digamma$ is a fairness query, and $\GAPair$ is the auxiliary information about the data collection process and the target population $\underlyingDist$.
\end{defn}

In this work, we formulate the problem of information incompleteness caused by selection bias by using the concept of {\em possible repairs} from Consistent Query Answering~\cite{bertossi2006consistent,dixit2021consistent}. This setup allows us to account for the uncertainty in the information about the target population $\underlyingDist$ and bound the estimates for the unfairness of the classifier on $\underlyingDist$. It is important to note that we do not actually compute the repairs themselves, but rather use the concept of possible repairs as a framework for addressing the incompleteness of information in the presence of selection bias.



\begin{defn}[Possible Repairs] Given a biased dataset $\bdata$, and auxiliary information $\GAPair=(\bcdag,\auxdata)$, we define the set of possible repairs of $\bdata$, denoted as $\repairs(\bdata)$, as the set of all datasets $D$ with the same schema as $\bdata$ such that: (1) $D \supsetneq \bdata$ and (2) $D$ is consistent with $\bcdag$ and $\auxdata$, i.e., it satisfies the constraints specified by these two components of $\GAPair$.
\label{def:possible-repair}
\end{defn}

\revb{Intuitively, any repair in $\repairs(\bdata)$ should be a superset of the biased dataset $\bdata$, adhering to the conditional independence constraints specified by $\bcdag$ and aligning with statistics derived from $\auxdata$ pertaining to the target distribution $\underlyingDist$.} As we may have no or partial information about $\underlyingDist$ in $\auxdata$, there are potentially infinite ways to repair $\bdata$. These constraints ensure that the repaired data aligns with any known information about $\underlyingDist$. Using this concept of possible repairs, we can now define the problem of consistently approximating a fairness query from biased data.



%
%
%

\label{section:recoverability}
\begin{figure*}
    \centering

{\scriptsize
\hspace{-3mm}\begin{tikzpicture}[node distance = 1.5cm, auto]
    \tikzstyle{initblock} = [rectangle, fill=initblockcolor, 
    text width=18em, text centered, rounded corners, minimum height=2em]
    \tikzstyle{propblock} = [rectangle, fill=propblockcolor, 
    text width=5em, text centered, rounded corners, minimum height=2em]
    \tikzstyle{decblock} = [rectangle, fill=decblockcolor, text width=5em, text centered, rounded corners, minimum height=2em]
    \tikzstyle{line} = [draw, -latex', thick]

    \node[inner sep=0pt] (diagram) at (-3.4,0)
    {\includegraphics[width=.68\textwidth]{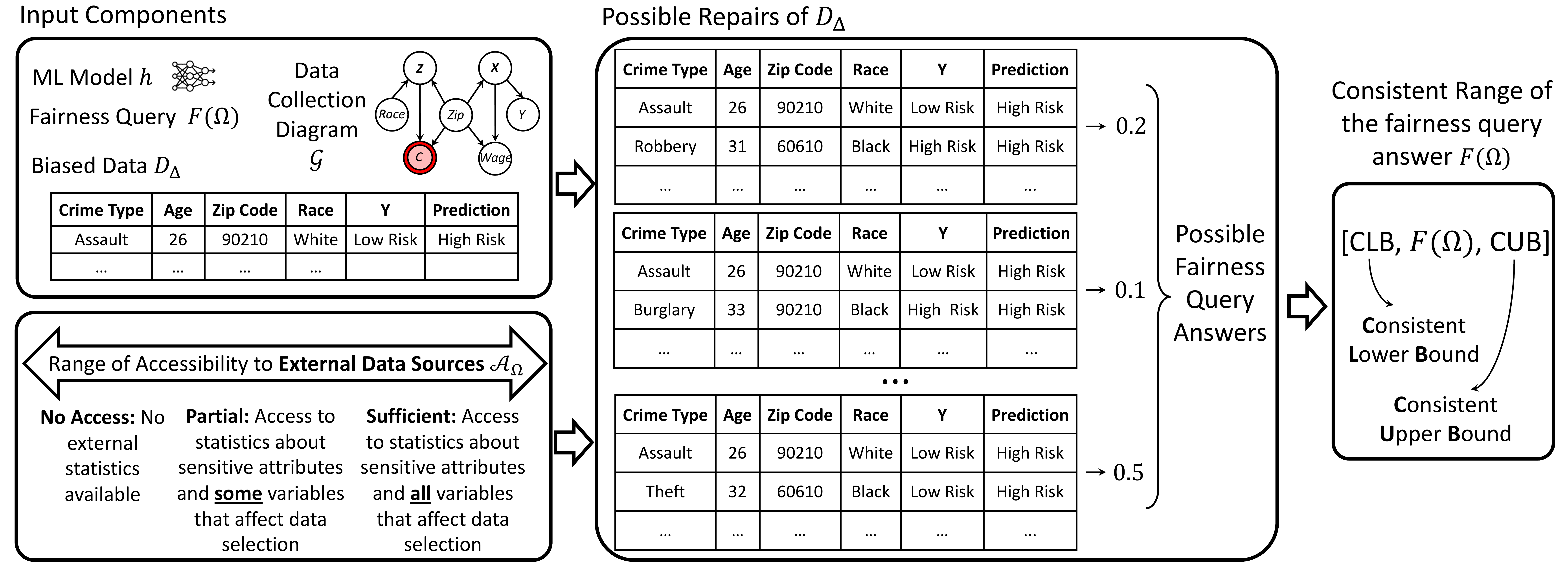}};
    \node [initblock] (init) at (5.9,1.5) {Identify $\Uvar$ in $\bcdag$ s.t. $(\Xvar \indep \SelectionVar \dsep \ProtectedAttr, \Admis, \Uvar)$};
    \node [decblock, below of=init, node distance=.8cm] (dec1) {$\ProtectedAttr$ available in $\auxdata$?};
    \node [decblock, right of=dec1, node distance=1.6cm] (dec2) {$\Admis$ available in $\auxdata$?};
    \node [propblock, left of=dec1, node distance=1.6cm] (prop1) {Prop.~\ref{proposition:no_external_data:dp:max_min_bound}};
    \node [decblock, below of=prop1, node distance=1.2cm] (dec3) {Selection depends on $\Admis$ \& $\Uvar$ available in $\auxdata$?};
    \node [decblock, below of=dec2, node distance=1.2cm] (dec4) {$\Uvar$ available in $\auxdata$?};
    \node [decblock, below of=dec1, node distance=1.2cm] (dec5) {Subset of $\Uvar$ available in $\auxdata$?};
    \node [propblock, below of=dec4, node distance=1.2cm] (prop2) {Prop.~\ref{prop:exactcal}};
    \node [propblock, below of=dec5, node distance=1.2cm] (prop3) {Prop.~\ref{proposition:limited_external_data:dp:max_min_bound}};
    \node [propblock, below of=dec3, node distance=1.2cm] (prop4) {Prop.~\ref{proposition:limited_external_data:dp:exactcalc}};
    \draw [line] (init) -- (dec1);
    \draw [line] (dec1) -- node [above]{no} (prop1);
    \draw [line] (dec1) -- node {yes} (dec2);
    \draw [line] ([xshift=1em]dec2.south) -- node {yes} ([xshift=1em]dec4.north);
    \draw [line] ([xshift=-1em]dec2.south) -|($([xshift=-1em]dec2.south)+(0, -0.35)$)|- node [sloped, anchor=center, left, yshift=1em]{no} ([yshift=2.3em]dec3.east);
    \draw [line] (dec3) -- node {no} (prop1);
    \draw [line] (dec4) -- node {no} (dec5);
    \draw [line] (dec5) -- node {yes} (prop3);
    \draw [line] (dec4) -- node {yes} (prop2);
    \draw [line] (dec3) -- node {yes} (prop4);
    \draw [line] (dec5) -- node [sloped, anchor=center, above, pos=0.5]{no} (prop1);

\end{tikzpicture}
}
    
    \vspace{-.2cm}
    \caption{\textmd{\revb{Left: An Illustration of the \cra\ process, which returns the range of possible answers $\underlyingDist$. The range is determined by computing the greatest lower bound and the lowest upper bound of the answer to the empirical fairness query over the set of possible repairs, providing a measure of the classifier's unfairness on the target population.} \revd{Right: Summary of scenarios where each proposition should be applied.}}}
    \vspace{-3mm}
    \label{fig:cra}
\end{figure*}

\begin{problem}[Consistent Range Approximation]
Given a \linebreak  fairness query application $\fairapp$, the problem of \ {\em  Consistent Range Approximation} (\cra) of a fairness query  $\digamma(\underlyingDist)$ is to determine the range of possible answers $[\lowerbound,\upperbound]$, where the $\lowerbound$ and $\upperbound$ denote the consistent lower and upper bounds of the 
answer to the empirical fairness query $\hat{\digamma}$ over possible repairs in $\repairs(\bdata)$, respectively.
Specifically,

\vspace{-5mm}
{\small
\begin{align*}
    \lowerbound = \min_{D\in \repairs(\bdata)} \hat{\digamma} (D), ~~~~\upperbound=\max_{D\in \repairs(\bdata)} \hat{\digamma} (D)
\end{align*}
}
\end{problem}

%
%
The set of possible repairs $\repairs(\bdata)$ includes all datasets that are consistent with the auxiliary information $\GAPair$, including a representative sample $D_{\underlyingDist}$ from the target distribution $\underlyingDist$. This means that the answer to ${\digamma}(\underlyingDist)$, which measures the fairness violation of a classifier on the target distribution, is guaranteed to be included within the range $[\lowerbound,\upperbound]$ (See figure~\ref{fig:cra}). This enables auditing the fairness of a classifier when the training data suffers from selection bias and limited information about the target distribution is available. One can also certify that a classifier is $\epsilon-$fair by checking that $\upperbound \leq \epsilon$. Additionally, as shown in Section~\ref{sec:recoverability:regularizer}, the upper bound can be used as a constraint or regularizer to train a classifier that is certifiably $\epsilon-$fair on the target population.
%
\par

Next, we investigate methods for obtaining consistent range answers to fairness queries in scenarios where the external data source is either absent or only provides partial information about the target population. While it is not possible to fully characterize the set of all repairs, we establish conditions under which closed-form solutions for consistent ranges can be obtained efficiently. 
\vspace{-0.2cm}

\subsubsection{Absence of External Data Sources}
\label{section:recoverability:withnoExternalData}
When there are no external data sources available, meaning $\auxdata=\emptyset$, we prove a sufficient condition for being able to estimate a consistent range $\sCondDemoParity$ using only the biased data $\bdata$. This range is represented by a closed-form formula, which makes it easy to calculate. In the context of algorithmic fairness, we're primarily interested in worst-case analysis, so we will focus on finding an upper bound for $\sCondDemoParity$ using this closed-form formula. Before we move forward, we'll also show a simple sufficient condition for $\sCondDemoParity=\sCondDemoParityb$, which means that the unfairness of a classifier on the target population can be obtained by computing the empirical fairness query on the biased data. \revb{Recall from the discussion below Definition~\ref{def:empfairquery} that we assume $\sCondDemoParityb=\seCondDemoParityDist{\dtrain}$, but we are still unable to directly evaluate $\sCondDemoParity$ with the biased training data $\dtrain$.}

%
%
\begin{proposition}
\label{prop:equality-actual-biased}
Given a fairness query application $\fairapp$, if the conditional independence ($\mbox{$\Xvar \indep \SelectionVar \mid \Admis, \ProtectedAttr$}$) holds,\linebreak then $\lowerbound = \sCondDemoParity=\sCondDemoParityb = \upperbound$.
\label{proposition:recoverability:with_external_data:demographic_parity}
\end{proposition}
%
\ifextended
\begin{proof} It is immediately implied by the independence $(\Xvar \indep \SelectionVar \mid \Admis, \ProtectedAttr)$ in the following steps. 
{\small \begin{flalign}
\indent& \begin{aligned}
  &\CondDemoParity{\Admis}\nonumber\\
  &= \frac{1}{\lvert\Admis\rvert}\sum_{\admis\in\Dom{\Admis}}\PrPos_{\underlyingDist}(h(\boldsymbol{x}) \mid \privG,\admis) - \PrPos_{\underlyingDist}(h(\boldsymbol{x}) \mid \protG,\admis) \nonumber\\
  &= \frac{1}{\lvert\Admis\rvert}\sum_{\admis\in\Dom{\Admis}}\PrPos_{\underlyingDist}(h(\boldsymbol{x}) \mid \privG,\admis,\SelectionVar=1) - \PrPos_{\underlyingDist}(h(\boldsymbol{x}) \mid\protG,\admis,\SelectionVar=1) \nonumber\\
  &=  \frac{1}{\lvert\Admis\rvert}\sum_{\admis\in\Dom{\Admis}}\PrPos_{\popul}(h(\boldsymbol{x}) \mid \privG,\admis) - \PrPos_{\popul}(h(\boldsymbol{x}) \mid \protG,\admis) \nonumber\\
  &= \CondDemoParityb{\Admis} \nonumber
\end{aligned}&&
\end{flalign}}
\end{proof}
\fi

\vspace{-0.2cm}
The independence condition in Proposition~\ref{prop:equality-actual-biased} does not hold in realistic situations because it requires the selection variable $\SelectionVar$ to only depend on the admissible attributes $\Admis$ and the protected attribute $\ProtectedAttr$, which is unrealistic. In most real-world scenarios, the selection variable is influenced by multiple factors, making it dependent on other variables as well. Therefore, in such situations, it's expected that $\sCondDemoParity \not =\sCondDemoParityb$. Next, we will establish a condition under which $\sCondDemoParity$ can be upper bounded using biased data.

\vspace{-0.1cm}
\begin{proposition}\label{proposition:no_external_data:dp:max_min_bound}
Given a fairness query application $\fairapp$, the following holds for any set of variables $\Uvar \subseteq \Allvar\cup\{Y\}$ such that $(\Xvar \indep \SelectionVar \dsep \ProtectedAttr, \Admis, \Uvar)$:
{\small
\begin{flalign}\label{eq:bound:no-external} \small
\indent&\begin{aligned} \small
    & 0\leq \sCondDemoParity\leq \upperbound\\
    & = \frac{1}{\lvert\Admis\rvert}\sum_{\admis\in\Dom{\Admis}} \big(\max_{\boldsymbol{u} \in \Dom{\boldsymbol{U}}}{ \PrPos_{\popul}(h(\boldsymbol{x})\mid\privG, \admis, \boldsymbol{u})}\\
    & - \min_{\boldsymbol{u} \in \Dom{\boldsymbol{U}}} {\PrPos_{\popul}(h(\boldsymbol{x})\mid\protG, \admis, \boldsymbol{u})}\big).
\end{aligned}&&
\end{flalign}
}
\end{proposition}

\ifextended
\begin{proof}
The proposition can be proved by 
upper bounding the expression $\PrPos_{\underlyingDist}(h(\boldsymbol{x})\mid \privG,\admis)$ and lower bounding  $\PrPos_{\underlyingDist}(h(\boldsymbol{x}) \mid \protG,\admis)$ for each $\admis\in\Admis$ in the definition of  $\CondDemoParity{\Admis}$ in Eq~\eqref{eq:binaryfairnessquery} using the independence assumption $(\Xvar \indep \SelectionVar \mid \Uvar, \ProtectedAttr, \Admis)$ as follows:

{\small
\begin{flalign}\label{eq:upperbound}
\indent&\begin{aligned}
    &\PrPos_{\underlyingDist}(h(\boldsymbol{x}) \mid \privG,\admis)\\
    &= \sum_{\boldsymbol{u} \in \Dom{\boldsymbol{U}}} \PrPos_{\underlyingDist}(h(\boldsymbol{x}) \mid \privG,\boldsymbol{u},\admis)\pr_{\underlyingDist}(\boldsymbol{u}\mid\privG,\admis)\\
    & = \sum_{\boldsymbol{u} \in \Dom{\boldsymbol{U}}} \PrPos_{\underlyingDist}(h(\boldsymbol{x}) \mid \privG,\boldsymbol{u},\admis,\SelectionVar=1)\ \pr_{\underlyingDist}(\boldsymbol{u}\mid\privG,\admis)\\
    &\leq \sum_{\boldsymbol{u} \in \Dom{\boldsymbol{U}}} (\max_{\boldsymbol{u^*} \in \Dom{\boldsymbol{U}}}\PrPos_{\underlyingDist}(h(\boldsymbol{x}) \mid \privG,\boldsymbol{u^*},\admis,\SelectionVar=1))\ \ \pr_{\underlyingDist}(\boldsymbol{u}\mid\privG,\admis)\\
    &= \max_{\boldsymbol{u^*} \in \Dom{\boldsymbol{U}}}\PrPos_{\underlyingDist}(h(\boldsymbol{x}) \mid \privG,\boldsymbol{u^*},\admis,\SelectionVar=1)\sum_{\boldsymbol{u} \in \Dom{\boldsymbol{U}}}\pr_{\underlyingDist}(\boldsymbol{u}\mid\privG,\admis)\\
    &= \max_{\boldsymbol{u^*} \in \Dom{\boldsymbol{U}}}\PrPos_{\underlyingDist}(h(\boldsymbol{x}) \mid \privG,\boldsymbol{u^*},\admis,\SelectionVar=1)\\
    &= \max_{\boldsymbol{u^*} \in \Dom{\boldsymbol{U}}}\PrPos_{\popul}(h(\boldsymbol{x}) \mid \privG,\boldsymbol{u^*},\admis)
\end{aligned}&&
\end{flalign}

\begin{flalign}\label{eq:lowerbound}
\indent&\begin{aligned}
    &\PrPos_{\underlyingDist}(h(\boldsymbol{x}) \mid \protG,\admis)\\
    &= \sum_{\boldsymbol{u} \in \Dom{\boldsymbol{U}}} \PrPos_{\underlyingDist}(h(\boldsymbol{x}) \mid \protG,\boldsymbol{u},\admis)\pr_{\underlyingDist}(\boldsymbol{u}\mid\protG,\admis)\\
    & = \sum_{\boldsymbol{u} \in \Dom{\boldsymbol{U}}} \PrPos_{\underlyingDist}(h(\boldsymbol{x}) \mid \protG,\boldsymbol{u},\admis,\SelectionVar=1)\pr_{\underlyingDist}(\boldsymbol{u}\mid\protG,\admis)\\
    &\geq \sum_{\boldsymbol{u} \in \Dom{\boldsymbol{U}}} (\min_{\boldsymbol{u^*} \in \Dom{\boldsymbol{U}}}\PrPos_{\underlyingDist}(h(\boldsymbol{x}) \mid \protG,\boldsymbol{u^*},\admis,\SelectionVar=1))\pr_{\underlyingDist}(\boldsymbol{u}\mid\protG,\admis)\\
    &= \min_{\boldsymbol{u^*} \in \Dom{\boldsymbol{U}}}\PrPos_{\underlyingDist}(h(\boldsymbol{x}) \mid \protG,\boldsymbol{u^*},\admis,\SelectionVar=1)\sum_{\boldsymbol{u}}\pr_{\underlyingDist}(\boldsymbol{u}\mid\protG,\admis)\\
    &= \min_{\boldsymbol{u^*} \in \Dom{\boldsymbol{U}}}\PrPos_{\underlyingDist}(h(\boldsymbol{x}) \mid \protG,\boldsymbol{u^*},\admis,\SelectionVar=1)\\
    &= \min_{\boldsymbol{u^*} \in \Dom{\boldsymbol{U}}}\PrPos_{\popul}(h(\boldsymbol{x}) \mid \protG,\boldsymbol{u^*},\admis)
\end{aligned}&&
\end{flalign}
}

Proposition~\ref{proposition:no_external_data:dp:max_min_bound} immediately follows from Eq~\eqref{eq:binaryfairnessquery}, ~\eqref{eq:upperbound} and ~\eqref{eq:lowerbound}.   
\end{proof}

\fi

The independence condition $(\Xvar \indep \SelectionVar \dsep \ProtectedAttr, \Admis, \Uvar)$ in Proposition~\ref{proposition:no_external_data:dp:max_min_bound} can be satisfied by setting $\boldsymbol{U}=\Pa(C)$, which forms the Markov Boundary of $C$ and ensures independence. Hence, Even without external data, an upper bound on $\sCondDemoParity$ can be computed using only the data collection diagram and biased training data.
\vspace{-0.2cm}
\subsubsection{Presence of External Data Sources}
\label{section:recoverability:withExternalData}

Now we establish conditions under which $\sCondDemoParity$ can be estimated or tightly bounded from biased data when we have access to external data sources that reveal varying levels of information about the target population.
\vspace{-0.1cm}
\begin{proposition}\label{prop:exactcal}
Given a fairness query application $\fairapp$, if there exists a set of variables $\mbox{$\Uvar \subseteq \Allvar\cup\{Y\}$}$ such\linebreak that $\mbox{$(\Xvar \indep \SelectionVar \dsep \ProtectedAttr, \Admis, \Uvar)$}$ holds, and where the auxiliary statistics $\Pr_{\underlyingDist}(\boldsymbol{u} \mid \protectedAttr, \admis)$, for all $u \in \Dom{\Uvar}$, $\admis \in \Dom{\Admis}$, and $\protectedAttr \in \{\protG, \privG\}$ can be obtained using external data sources $\auxdata$, then the fairness query $\fairnessquery(\underlyingDist)$ can be computed using the following equation:
{\small
\begin{flalign}\small
\label{eq:exactcalc}\indent&\begin{aligned}
    &\lowerbound = \fairnessquery(\underlyingDist) = \upperbound \\
    &= \frac{1}{\lvert\Admis\rvert}\sum_{\admis\in\Dom{\Admis}}\sum_{\boldsymbol{u}\in\Dom{\Uvar}} \big(\PrPos_{\popul}(h(\boldsymbol{x}) \mid \privG, \admis, \boldsymbol{u}) \  \pr_{\underlyingDist}(\boldsymbol{u} \mid \privG, \admis) \\
    &- \PrPos_{\popul}(h(\boldsymbol{x}) \mid \protG, \admis, \boldsymbol{u}) \ \pr_{\underlyingDist} (\boldsymbol{u} \mid \protG, \admis)\big)
\end{aligned}&&
\end{flalign} 
}
\end{proposition}

%
%
\ifextended
\begin{proof}  The following equations are obtained from
the law of total expectation, the independence $(\Xvar \indep \SelectionVar \mid \ProtectedAttr, \Uvar,\Admis)$. For each $\protectedAttr \in {\protG, \privG}$ and $\admis\in\Admis$, it holds that:

{\small
\begin{flalign}\label{eq:cdp1}
\indent&\begin{aligned}
    &\PrPos_{\underlyingDist}(h(\boldsymbol{x})\mid \protectedAttr, \admis)\\
    &= \sum_{\boldsymbol{u} \in \Dom{\boldsymbol{U}}}\PrPos_{\underlyingDist}(h(\boldsymbol{x})\mid \protectedAttr, \boldsymbol{u}, \admis)\ \pr_{\underlyingDist}(\boldsymbol{u}\mid\protectedAttr, \admis) \\
    &= \sum_{\boldsymbol{u} \in \Dom{\boldsymbol{U}}}\PrPos_{\underlyingDist}(h(\boldsymbol{x})\mid\protectedAttr, \boldsymbol{u}, \admis, \SelectionVar)\ \pr_{\underlyingDist}(\boldsymbol{u}\mid\protectedAttr, \admis)\\
    &= \sum_{\boldsymbol{u} \in \Dom{\boldsymbol{U}}}\PrPos_{\popul}(h(\boldsymbol{x})\mid\protectedAttr, \boldsymbol{u}, \admis) \ \pr_{\underlyingDist}(\boldsymbol{u}\mid\protectedAttr, \admis)
\end{aligned}&&
\end{flalign}
}

Eq~\eqref{eq:exactcalc} immediately follows from the preceding  equations.
\end{proof}
\fi

Eq~\eqref{eq:exactcalc} calculates {$\mbox{$\PrPos_{\popul}(h(\boldsymbol{x})\mid\protectedAttr, \admis, \boldsymbol{u})$}$}, which can be estimated from biased data. However, $\auxdata$ must be used to compute { $\mbox{$\pr_{\underlyingDist}(\boldsymbol{u}\mid\protectedAttr, \admis)$}$}.
The independence assumption $(\Xvar \indep \SelectionVar \dsep \ProtectedAttr, \Admis, \Uvar)$ is crucial to apply Proposition~\ref{proposition:no_external_data:dp:max_min_bound} and \ref{prop:exactcal}. By setting $\boldsymbol{U}=\pa(C)$, i.e., the parents of the selection variable $\SelectionVar$ in the data collection diagram $\bcdag$, we can always find variables that satisfy this assumption (cf. Section~\ref{section:recoverability:withnoExternalData}). Therefore, these propositions are applicable even when the entire data collection diagram $\bcdag$ is unknown and we have access to information only about $\pa(C)$.

However, when the data collection diagram $\bcdag$ is available, we can select a minimal set of variables $\boldsymbol{U}$ that satisfies the independence condition and for which auxiliary information in $\auxdata$ is available. This facilitates applying the proposition in settings where $\auxdata$ does not contain information about the entire set of variables in $\pa(C)$. Specifically, whenever the selection variable depends on the training label $Y$, i.e., $Y \in \Pa(C)$, using the data collection diagram $\bcdag$ enables identifying a set of variables $\boldsymbol{U}$ that does not contain $Y$ and still satisfy the independence condition. This enables the use of Proposition~\ref{prop:exactcal} even if no information about the label $Y$ is available in $\auxdata$, which is often the case in practice. We will show next that such a set of variables always exists, hence one can establish an upper bound for a fairness notion in general, even if no external data source about the outcome variable $Y$ is required.
\vspace{-0.2cm}
\begin{proposition} 
\label{prop:Uindepdance}
Given a data collection diagram $\bcdag$, there always exists a set of variables $\Uvar \subseteq \Pa(\SelectionVar) \cup MB(\Yvar) \setminus \{\SelectionVar, Y\}$ that satisfy the conditional independence $(\Xvar \indep \SelectionVar \dsep \ProtectedAttr, \Admis, \Uvar)$. (Note that $\Pa(\SelectionVar)$ represents the parents of $\SelectionVar$ in $\bcdag$, and $MB(\Yvar)$ denotes the Markov Boundary of $Y$ in $\bcdag$.)


\end{proposition}

\vspace{-0.3cm}
\begin{example}
In the data collection diagram in~\Cref{fig:police-dag:b}, the conditional independence $\mbox{$(\Xvar \indep \SelectionVar \dsep \ProtectedAttr, \Admis, \Uvar)$}$ holds for $\Uvar=\pa(C)=\{Z,ZIPCode\}$. This means that $\sCondDemoParity$ can be estimated from a biased sample collected according to the diagram in \Cref{fig:police-dag:b} and using auxiliary information to estimate statistics of the form $\mbox{$\pr_{\underlyingDist}(\boldsymbol{u} \mid s, \admis)$}$.  Now consider a similar scenario, except that the selection variable is a function of $Y$ and $ZIPCode$, i.e., $\pa(C)=\{Y,ZIPCode\}$. In this case, the independence condition holds for $\Uvar=\pa(C)=\{Y,ZIPCode\}$. However, if information about the label $Y$ is not available in external data sources, one can select $\boldsymbol{U}= \{ZIPCode, \boldsymbol{W}\}$ that satisfy the independence condition $(\Xvar \indep \SelectionVar \dsep \ProtectedAttr, \Admis, \Uvar)$ and use Proposition~\ref{prop:exactcal} to compute $\fairnessquery(\underlyingDist)$.
\end{example}

Propositions~\ref{proposition:no_external_data:dp:max_min_bound} and~\ref{prop:exactcal} examine different ends of the spectrum in terms of the availability of external data. The former requires no external data, while the latter requires sufficient external data for exact computation of a fairness query. In practice, one may have access to a level of external data that falls in between these two extremes. In such cases, it is important to note that the selection variable, $\SelectionVar$, may depend on a high-dimensional set of variables, and thus the set of variables $\Uvar$ for which the conditions of Proposition~\ref{proposition:no_external_data:dp:max_min_bound} hold could also consist of a high-dimensional set of variables. This may make it infeasible to collect auxiliary information for computing all the statistics $\Pr_{\underlyingDist}(\boldsymbol{u} \mid \protectedAttr, \admis)$ needed for exact computation. Thus, in this case, we investigate the middle of the spectrum, where some auxiliary information about the target population is available but not enough for applying \Cref{prop:exactcal}. We show that this limited amount of auxiliary information can be used to compute a tighter upper bound for $\sCondDemoParity$ than that established in Proposition~\ref{proposition:no_external_data:dp:max_min_bound}. Specifically, we consider similar assumptions as in Proposition~\ref{prop:exactcal}, but in situations where external data sources have only partial information about the statistics $\mbox{$\pr_{\underlyingDist} (\boldsymbol{u} \mid \protG, \admis)$}$.

\vspace{-0.2cm}
\begin{proposition}\label{proposition:limited_external_data:dp:max_min_bound}
Given a fairness query application $\fairapp$ and a subset of variables $\Uvar \subseteq \Allvar\cup\{Y\}$ such that $(\Xvar \indep \SelectionVar \dsep \ProtectedAttr, \Admis, \Uvar)$ holds in $\bcdag$, if $\auxdata$ only permit computation of the auxiliary statistics $\pr_{\underlyingDist}(\boldsymbol{u}'\mid\protectedAttr,\admis)$ for all $\boldsymbol{u}' \in \Dom{\boldsymbol{U}'}$ and $ \protectedAttr \in \{\protG, \privG\}$ for some subset $\boldsymbol{U}' \subsetneq \boldsymbol{U}$, then the following upper bound can be computed for $\sCondDemoParity$:
{\small
\begin{flalign}
    \indent&\begin{aligned}
    &0\leq \sCondDemoParity
    \leq \upperbound = \frac{1}{\lvert\Admis\rvert}\sum_{\admis\in\Dom{\Admis}} \sum_{\boldsymbol{u}' \in \boldsymbol{U}'}\\
    &\Big( \pr_{\underlyingDist}(\boldsymbol{u}'\mid\privG, \admis)\max_{\boldsymbol{u}^*\in \Dom{\Uvar\setminus\boldsymbol{U}'}}(\PrPos_{\popul}(h(\boldsymbol{x})\mid\privG, \admis, \boldsymbol{u}', \boldsymbol{u}^*))\\
    &- \pr_{\underlyingDist}(\boldsymbol{u}'\mid\protG, \admis)\min_{\boldsymbol{u}^*\in \Dom{\Uvar\setminus\boldsymbol{U}'}}(\PrPos_{\popul}(h(\boldsymbol{x})\mid\protG, \admis, \boldsymbol{u}', \boldsymbol{u}^*))\Big)\label{eq:tightbound}
\end{aligned}&
\end{flalign}
}

\end{proposition}
\ifextended
\begin{proof}
Let's partition $\boldsymbol{U}$ into $\boldsymbol{U}'$ and $\boldsymbol{U}\setminus\boldsymbol{U}'$ in Eq~\eqref{eq:exactcalc} the following inequality obtained from applying the Fréchet inequality to $\pr_{\underlyingDist}(\boldsymbol{u}', \boldsymbol{u}^* \mid \protectedAttr, \admis)$:

{\small
\begin{flalign}
\indent&
\begin{aligned}
&\CondDemoParity{\Admis}= \frac{1}{\lvert\Admis\rvert}\sum_{\admis\in\Dom{\Admis}}\sum_{\boldsymbol{u}'\in\Dom{U'}}\sum_{\boldsymbol{u}^*\in \Dom{\Uvar\setminus\boldsymbol{U}'}}\\
&\Big(\PrPos_{\popul}(h(\boldsymbol{x}) \mid \privG, \boldsymbol{u}', \boldsymbol{u}^*, \admis) \  \pr_{\underlyingDist}(\boldsymbol{u}', \boldsymbol{u}^* \mid \privG, \admis) \\
&- \PrPos_{\popul}(h(\boldsymbol{x}) \mid \protG, \boldsymbol{u}', \boldsymbol{u}^*, \admis) \ \pr_{\underlyingDist} (\boldsymbol{u}', \boldsymbol{u}^* \mid \protG, \admis)\Big)\\
&\leq \frac{1}{\lvert\Admis\rvert}\sum_{\admis\in\Dom{\Admis}}\sum_{\boldsymbol{u}'\in\Dom{U'}}\sum_{\boldsymbol{u}^*\in \Dom{\Uvar\setminus\boldsymbol{U}'}}\\
&\Big( \PrPos_{\popul}(h(\boldsymbol{x}) \mid \privG, \boldsymbol{u}', \boldsymbol{u}^*, \admis) \  \pr_{\underlyingDist}(\boldsymbol{u}'\mid \privG, \admis) \\
&- \PrPos_{\popul}(h(\boldsymbol{x}) \mid \protG, \boldsymbol{u}', \boldsymbol{u}^*, \admis) \ \pr_{\underlyingDist} (\boldsymbol{u}' \mid \protG, \admis)\Big) \label{eq:midineq}
\end{aligned}&&
\end{flalign}
}

Then we upper bound the expression $\PrPos_{\popul}(h(\boldsymbol{x}) \mid \privG, \boldsymbol{u}', \boldsymbol{u}^*, \admis)$ and lower bound  $\PrPos_{\popul}(h(\boldsymbol{x}) \mid \protG, \boldsymbol{u}', \boldsymbol{u}^*, \admis)$ in Eq~\eqref{eq:midineq} using the independence $(\Xvar \indep \SelectionVar \mid \boldsymbol{U}, \ProtectedAttr, \Admis)$ in as similar steps as in the proof of \cref{proposition:no_external_data:dp:max_min_bound}:
{\small
\begin{flalign}\label{eq:tightupperbound}
\indent&\begin{aligned}
\PrPos_{\popul}(h(\boldsymbol{x}) \mid \privG, \boldsymbol{u}', \boldsymbol{u}^*, \admis)\leq \max_{\boldsymbol{u}^*\in \Dom{\Uvar\setminus\boldsymbol{U}'}}(\PrPos_{\popul}(h(\boldsymbol{x})\mid\privG, \boldsymbol{u}, \admis))
\end{aligned}&&   
\end{flalign}
}
{\small
\begin{flalign}
\indent&\begin{aligned}
\PrPos_{\popul}(h(\boldsymbol{x}) \mid \protG, \boldsymbol{u}', \boldsymbol{u}^*, \admis)\geq \min_{\boldsymbol{u}^*\in \Dom{\Uvar\setminus\boldsymbol{U}'}}(\PrPos_{\popul}(h(\boldsymbol{x})\mid\protG, \boldsymbol{u}, \admis))
\end{aligned}&&
\label{eq:tightlowerbound}
\end{flalign}
}

Eq~\eqref{eq:tightbound} immediately follows from Eq~\eqref{eq:midineq}, ~\eqref{eq:tightupperbound} and ~\eqref{eq:tightlowerbound}.

\end{proof}
\fi

\par

Now we consider a scenario where external data source about the entire set of admissible variables $\Admis$ is unavailable, which can be a challenge when dealing with fairness definitions based on error rate balance, such as equality of odds, where the training label $Y$ is included in $\Admis$. \reva{In these cases, Propositions~\ref{proposition:no_external_data:dp:max_min_bound} and~\ref{proposition:limited_external_data:dp:max_min_bound} can only be applied in the presence of auxiliary information about $Y$, which is difficult to acquire in practice.} However, we show that under certain assumptions about the data collection process, it is still possible to bound $\sCondDemoParity$ from biased data even in the absence of auxiliary information about $\Admis$. 

\begin{proposition}\label{proposition:limited_external_data:dp:exactcalc}
Let $\fairapp$ be a fairness query application and $\bcdag$ the corresponding data collection diagram. If $\Admis\cap\Pa(\SelectionVar)=\emptyset$ in $\bcdag$, i.e., data selection does not directly depend on the admissible variables $\Admis$, then for a set of variables $\Uvar \subseteq \Allvar\cup\{Y\}$ that satisfies the conditional independence $(\Xvar \indep \SelectionVar \dsep \ProtectedAttr, \Admis, \Uvar)$, if the target population statistics $\pr_{\underlyingDist}(\protectedAttr,\boldsymbol{u})$ for all $\boldsymbol{u} \in \Dom{\boldsymbol{U}}$ and $ \protectedAttr \in \{\protG, \privG\}$ can be computed from external data sources $\auxdata$, then $\sCondDemoParity$ can be computed using the following formula:
{\small
\begin{flalign}
\indent&\begin{aligned}
& \lowerbound = \sCondDemoParity = \upperbound\\
& = \frac{1}{\lvert\Admis\rvert}\sum_{\boldsymbol{u}\in\Dom{\boldsymbol{U}}} \big(\PrPos_{\popul}(h(\boldsymbol{x}) \mid \privG, \admis, \boldsymbol{u}) \  w(\privG, \boldsymbol{u}, \admis) 
\\
& -  \PrPos_{\popul}(h(\boldsymbol{x}) \mid \protG, \admis, \boldsymbol{u}) \   w(\protG, \boldsymbol{u}, \admis)\big).\label{eq:exactcalc:missA}
\end{aligned}&
\end{flalign}
}
\noindent where $w(\boldsymbol{u}, \protectedAttr, \admis)=\frac{\sum_{\boldsymbol{x} \in \Dom{\Xvar}}{\pr_{\popul}(\admis,\boldsymbol{x}\mid s,\boldsymbol{u})\pr_{\underlyingDist}( s,\boldsymbol{u})}}{\sum_{\boldsymbol{u}^* \in \Dom{\boldsymbol{U}}}{\sum_{\boldsymbol{x} \in \Dom{\Xvar}}{\pr_{\popul}(\admis,\boldsymbol{x}\mid s,\boldsymbol{u^*})\pr_{\underlyingDist}( s,\boldsymbol{u^*})}}}$ 
\end{proposition}
%

\ifextended
\begin{proof}
Since $\Admis \cap\Pa(\SelectionVar)=\emptyset$, the conditional independence $\Admis,\Xvar \indep \SelectionVar \mid \ProtectedAttr,\Uvar$ holds for $\Uvar=\Pa(\SelectionVar)$ such that $\Admis \cap \Uvar=\emptyset$. The following equations are obtained from the above the independence assumption, law of total probability, the Bayes' Theorem and marginalization on $\Xvar$. For each $\protectedAttr \in \{\protG, \privG\}$, we can estimate $\pr_{\underlyingDist}(\boldsymbol{u}\mid s,\admis)$ in Eq~\eqref{eq:cdp1} as follows:
{\small
\begin{flalign}
\indent&\begin{aligned}
    & \pr_{\underlyingDist}(\boldsymbol{u}\mid s,\admis)
    =\frac{\sum_{x\in\Dom{\Xvar}}{\pr_{\underlyingDist}(\admis,\boldsymbol{x}\mid s,\boldsymbol{u})\pr_{\underlyingDist}(s,\boldsymbol{u})}}{\sum_{x\in\Dom{\Xvar}}{\pr_{\underlyingDist}(\admis,\boldsymbol{x}\mid s)\pr_{\underlyingDist}(s)}}\\
    &=\frac{\sum_{x\in\Dom{\Xvar}}{\pr_{\underlyingDist}(\admis,\boldsymbol{x}\mid s,\boldsymbol{u})\pr_{\underlyingDist}( s,\boldsymbol{u})}}{\sum_{u^*\in\Dom{\Uvar}}{\sum_{x\in\Dom{\Xvar}}{\pr_{\underlyingDist}(\admis,\boldsymbol{x}\mid s,\boldsymbol{u^*})\pr_{\underlyingDist}( s,\boldsymbol{u^*})}}} \\
    &=\frac{\sum_{x\in\Dom{\Xvar}}{\pr_{\underlyingDist}(\admis,\boldsymbol{x}\mid s,\boldsymbol{u},\SelectionVar=1)\pr_{\underlyingDist}( s,\boldsymbol{u})}}{\sum_{u^*\in\Dom{\Uvar}}{\sum_{x\in\Dom{\Xvar}}{\pr_{\underlyingDist}(\admis,\boldsymbol{x}\mid s,\boldsymbol{u^*},\SelectionVar=1)\pr_{\underlyingDist}( s,\boldsymbol{u^*})}}} \\
    &=\frac{\sum_{x\in\Dom{\Xvar}}{\pr_{\popul}(\admis,\boldsymbol{x}\mid s,\boldsymbol{u})\pr_{\underlyingDist}( s,\boldsymbol{u})}}{\sum_{u^*\in\Dom{\Uvar}}{\sum_{x\in\Dom{\Xvar}}{\pr_{\popul}(\admis,\boldsymbol{x}\mid s,\boldsymbol{u^*})\pr_{\underlyingDist}( s,\boldsymbol{u^*})}}}
\end{aligned}&&\label{eq:estimate-missing-a}
\end{flalign}
}

Eq~\eqref{proposition:limited_external_data:dp:exactcalc} immediately follows from Eq~\eqref{eq:binaryfairnessquery},~\eqref{eq:cdp1}, and~\eqref{eq:estimate-missing-a}.
\end{proof}
\fi
\revd{
\subsubsection{Summary of the results for the \cra\ of fairness query}
In this section, we've introduced methods to address the Consistent Range Answer (CRA) of fairness queries $\digamma(\Omega)$ under various scenarios, depending on the availability of an external data source ($A_\Omega$) and the requirements of the data collection diagram ($\bcdag$) (refer to Figure~\ref{fig:cra}). Without an external data source, we depend solely on the data collection diagram and use Proposition~\ref{proposition:no_external_data:dp:max_min_bound} to estimate the CUB of $\sCondDemoParity$. When external data sources are available, providing statistics about admissible variables, sensitive attributes, and a variable set $\Uvar$ that meets the conditional independence constraint, we use Proposition~\ref{prop:exactcal} to directly approximate $\sCondDemoParity$ (CUB=CLB). If only a subset of $\Uvar$ is available from the external data source, we apply Proposition~\ref{proposition:limited_external_data:dp:max_min_bound} to derive a tighter CUB using the available statistics. If the selection process doesn't depend on any admissible attribute, we only need external data sources about $\Uvar$ and sensitive attributes to approximate $\sCondDemoParity$ using Proposition~\ref{proposition:limited_external_data:dp:exactcalc} . 
}
\ignore{ 
\revd{
\subsubsection{Summary of the result for \cra of fairness query}\label{section:cra:summary}

In this section, we have proposed solutions for the Consistent Range Answer (CRA) of fairness queries $\digamma(\Omega)$ under diverse scenarios, contingent on the accessibility of an external data source ($A_\Omega$) and the requirement of the data collection diagram ($G$). The scenarios we examined include (see the flowchart in Figure~\ref{fig:cra}):
\begin{enumerate}
\item If no external data source is available, we rely solely on the data collection process. Using Proposition~\ref{proposition:no_external_data:dp:max_min_bound}, we estimate the CUB of $\sCondDemoParity$ based on the data collection diagram.

\item When external data sources are accessible and provide statistics about admissible variables, sensitive attributes and a variable set $\Uvar$ that satisfy the conditional independence constraint (which is always possible by setting parents of the selection variable as $\Uvar$), we use Proposition~\ref{prop:exactcal} to directly approximate $\sCondDemoParity$ (CUB=CLB).

\item In the same situation as (2), except that instead of the entire $\Uvar$, only a subset of it is available in the external data source, we apply Proposition~\ref{proposition:limited_external_data:dp:max_min_bound}. This proposition allows us to derive a tighter CUB by incorporating the available statistics.
\babak{propagate}
\item In the scenario where the selection process does not depend on any of the admissible attribute, we only require external data sources about $\Uvar$ and Sensitive attributes to approximate . By utilizing these specific statistics, we can approximate $\sCondDemoParity$ by applying Proposition~\ref{proposition:limited_external_data:dp:exactcalc}.
\end{enumerate}
By considering these diverse scenarios, we have addressed the challenge of information incompleteness and provided solutions to obtain consistent ranges of for fairness query answers. 
}
}

\ignore{
\revd{
\subsubsection{Summary} 
In this section, we have proposed solutions for the \cra for fairness query $\sCondDemoParity$ under diverse scenarios, contingent on the accessibility of an external data source ($\auxdata$), and the requirement of the data collection diagram ($\bcdag$). The scenarios we examined include:
\begin{enumerate} 
\item When there is no access to any external data source, i.e., $\auxdata=\emptyset$, we use Proposition~\ref{proposition:no_external_data:dp:max_min_bound} to determine the CUB of $\sCondDemoParity$.
\item When we have external data sources that allow computation of the statistics $\pr_{\underlyingDist} (\boldsymbol{u} \mid \protG, \admis)$, where the condition $(\Xvar \indep \SelectionVar \dsep \ProtectedAttr, \Admis, \Uvar)$ holds in $\bcdag$, we utilize Proposition~\ref{prop:exactcal} to determine the CUB of $\sCondDemoParity$.
\item In cases where the available external data source is inadequate for computing $\pr_{\underlyingDist} (\boldsymbol{u} \mid \protG, \admis)$, but can be used to compute $\pr_{\underlyingDist} (\boldsymbol{u'} \mid \protG, \admis)$ for some $\boldsymbol{U'}\subsetneq\Uvar$, we apply Proposition~\ref{proposition:limited_external_data:dp:max_min_bound} to derive a tighter CUB than that obtained using Proposition~\ref{proposition:no_external_data:dp:max_min_bound}.
\item When the condition $\Admis \ \cap \ \Pa(\SelectionVar)=\emptyset$ is satisfied in $\bcdag$, we only need external data sources that can compute $\pr_{\underlyingDist} (\boldsymbol{u'}, \protectedAttr)$ to approximate $\sCondDemoParity$.
\end{enumerate}
}
}

\vspace{-3mm}
\subsection{Fair ML with \cra}\label{sec:recoverability:regularizer}


The results established for \cra\ of $\fairnessquery(\underlyingDist)$ from biased data can be used to train predictive models by solving the following constrained risk minimization problem (cf. Section~\ref{section:background}), which enforces an upper bound on $\fairnessquery(\underlyingDist)$ obtained via \cra :


\begin{equation}\label{eq:primal}
\begin{aligned}
    \min_{\classifier \in \mc{H}}  \EX_{\popul}[L(\classifier(\datap), y)]  \quad  \text{s.t.} ~~\upperbound(\fairnessquery(\underlyingDist)) \leq \tau
\end{aligned}
\end{equation}
where $\upperbound(\fairnessquery(\underlyingDist))$ is the consistent upper bound of a fairness query 
that can be computed from biased training data via the results established in Propositions~\ref{proposition:no_external_data:dp:max_min_bound} to~\ref{proposition:limited_external_data:dp:exactcalc}, and $\tau$ is a threshold that can be tuned to trade fairness with accuracy based on application scenarios. However, directly solving this constrained optimization problem for an arbitrary ML model is difficult. Instead, we can adopt the idea of the penalty method~\cite{nocedal1999numerical,smith1997penalty,berk2017convex, kamishima2012fairness,bechavod2017learning,zhang2021omnifair} and convert the constrained optimization problem in Eq~\eqref{eq:primal} to an unconstrained problem by adding a regularization term that penalizes high upper bound of unfairness to the objective function:
\begin{equation}
\begin{aligned}
    \min_{\classifier \in \mc{H}}  \EX_{\popul}[L(\classifier(\datap), y)]+\lambda\cdot \upperbound(\fairnessquery(\underlyingDist)) ~\label{eq:reqopt}
\end{aligned}
\end{equation}

We use Algorithm~\ref{alg:fair_training} to solve the optimization problem in Eq~\eqref{eq:reqopt}. To balance fairness and accuracy, we introduce a fairness threshold $\tau$ and use $\max\big(\upperbound(\fairnessquery(\underlyingDist)), \tau\big)$ in the fairness constraint of the optimization problem. This allows for adjusting the trade-off between fairness and accuracy based on the specific requirements of the application. Additionally, since $\fairnessquery(\underlyingDist)$ is not differentiable for a binary classifier, we use a differentiable relaxation based on the classifier's output probabilities rather than decisions.
In each iteration, the algorithm computes the consistent upper bound $\upperbound$ of the classifier's unfairness using the proposed \cra\ framework, considering the availability of the external data source $\auxdata$. 

\vspace{-0.2cm}
{ \small
\begin{algorithm}
\caption{\small Training a fair ML model from biased data}
\label{alg:fair_training}
\begin{algorithmic}[1]
\Input Biased training dataset $\dtrain$, auxiliary information $\GAPair$, fairness query $\fairnessquery$, unfairness threshold $\tau$, penalty coefficient $\lambda$, learning rate $\eta$.
\Output Fair ML model $\classifier_\theta$ with parameter $\theta$.

\State $\classifier_\theta \gets \text{random\_initialization}$

\While{not converged}
    \State $loss \gets$ empirical\_loss($\classifier,\dtrain$)
    \State $\upperbound \gets$ CRA($h_\theta,\digamma,\dtrain,\GAPair$) \Comment results in Section~\ref{sec:cra}
    \State $loss \gets loss+\lambda\cdot\max\{\upperbound,\tau\}$
    \State $\theta \gets \theta$ - $\eta\cdot$ gradient($loss$, $\theta$) 
\EndWhile

\end{algorithmic}
\end{algorithm}
}

%% file: Section_Tex_Files/conditions.tex
\section{Fairness and Selection Bias}
\label{section:introducing_bias_out_of_the_blue}

\begin{figure}
     \centering
     \begin{subfigure}[b]{0.23\linewidth}
         \centering
         \includegraphics[width=\textwidth]{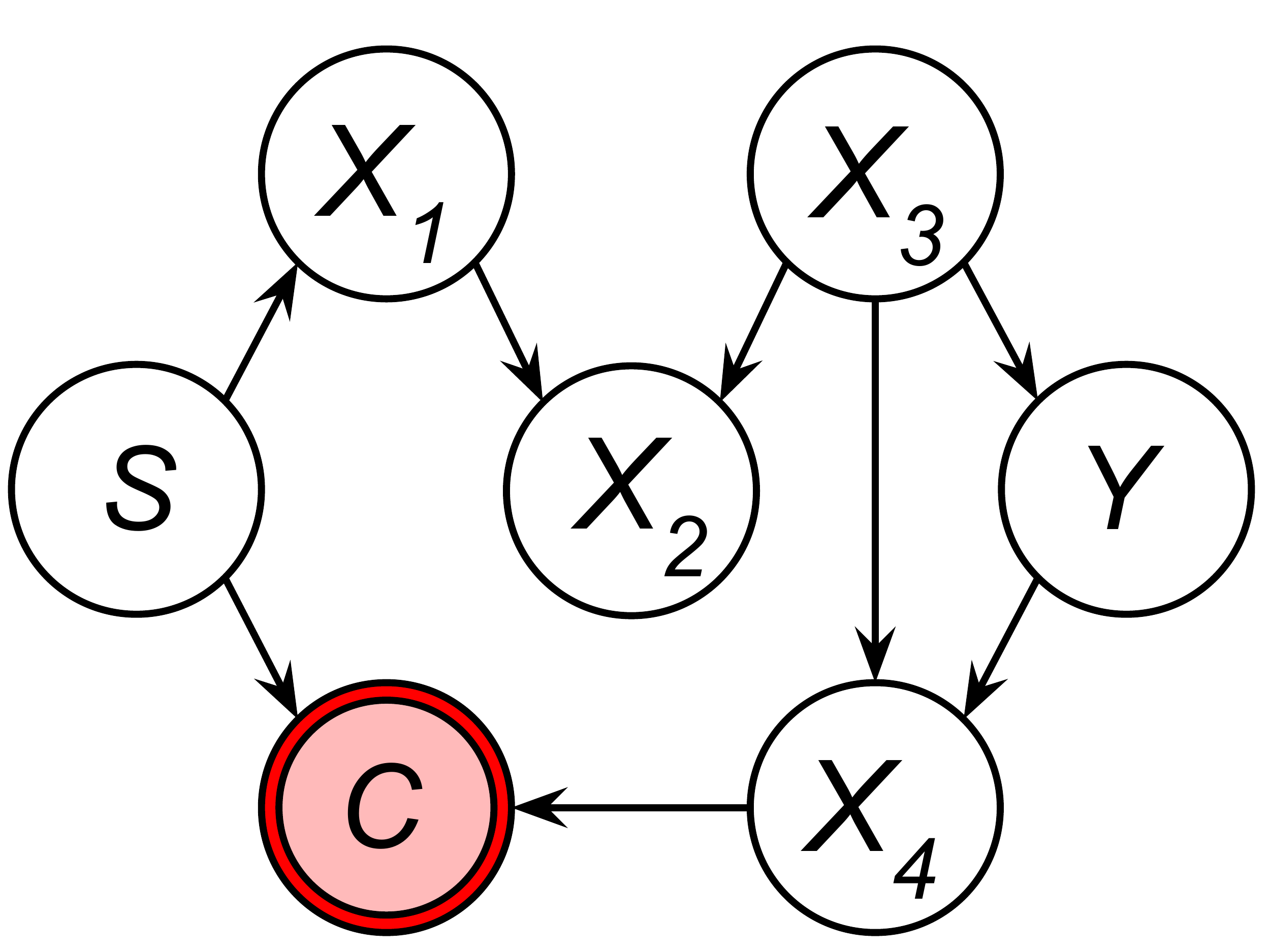}
         \caption{}
         \label{figure:structural_conditions:based_on_S_X4}
     \end{subfigure}
     \hspace{.07cm}
     \begin{subfigure}[b]{0.23\linewidth}
         \centering
         \includegraphics[width=\textwidth]{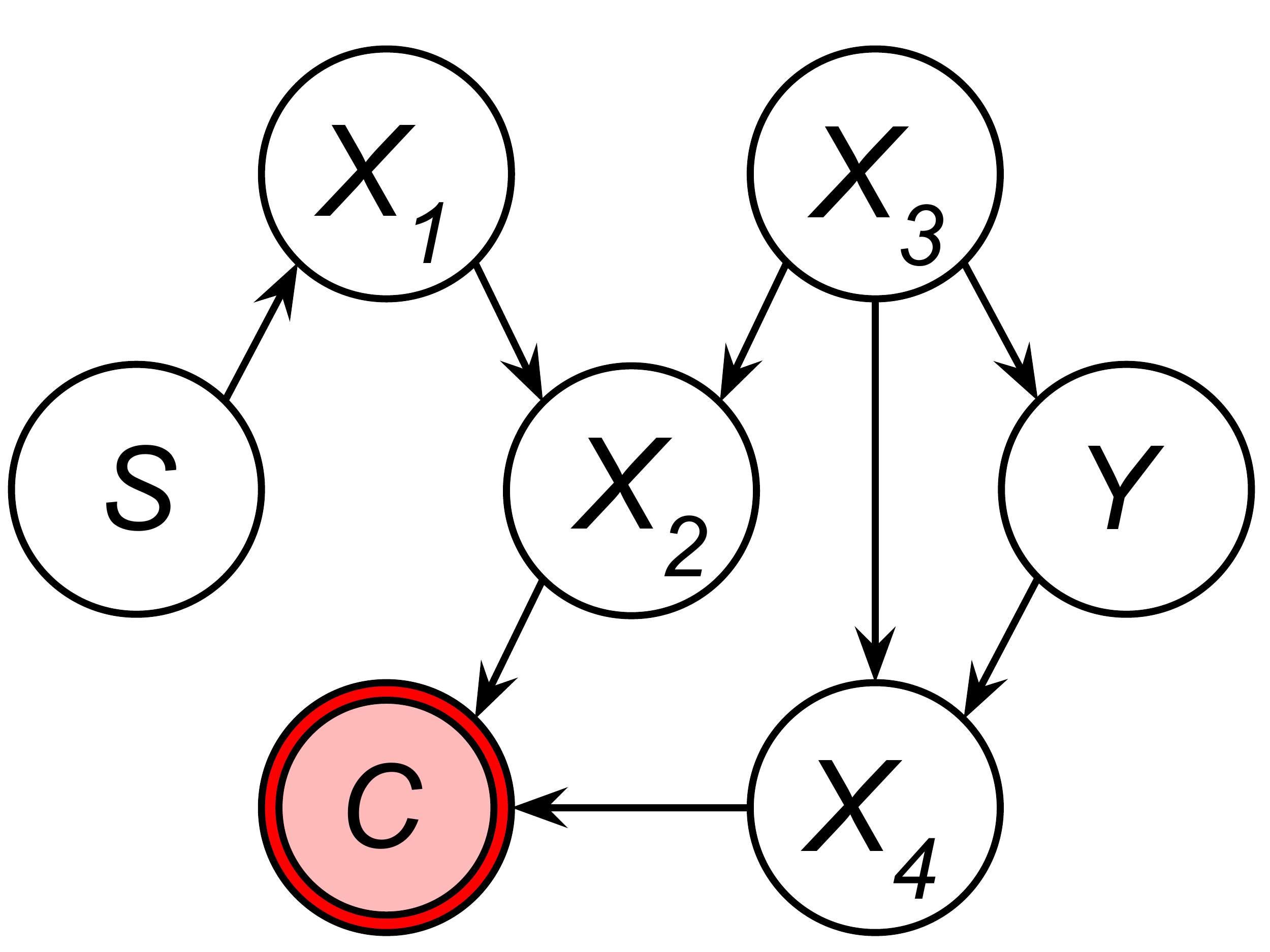}
         \caption{}
         \label{figure:structural_conditions:based_on_X2_X4}
     \end{subfigure}
    \hspace{.07cm}
     \begin{subfigure}[b]{0.23\linewidth}
         \centering
         \includegraphics[width=\textwidth]{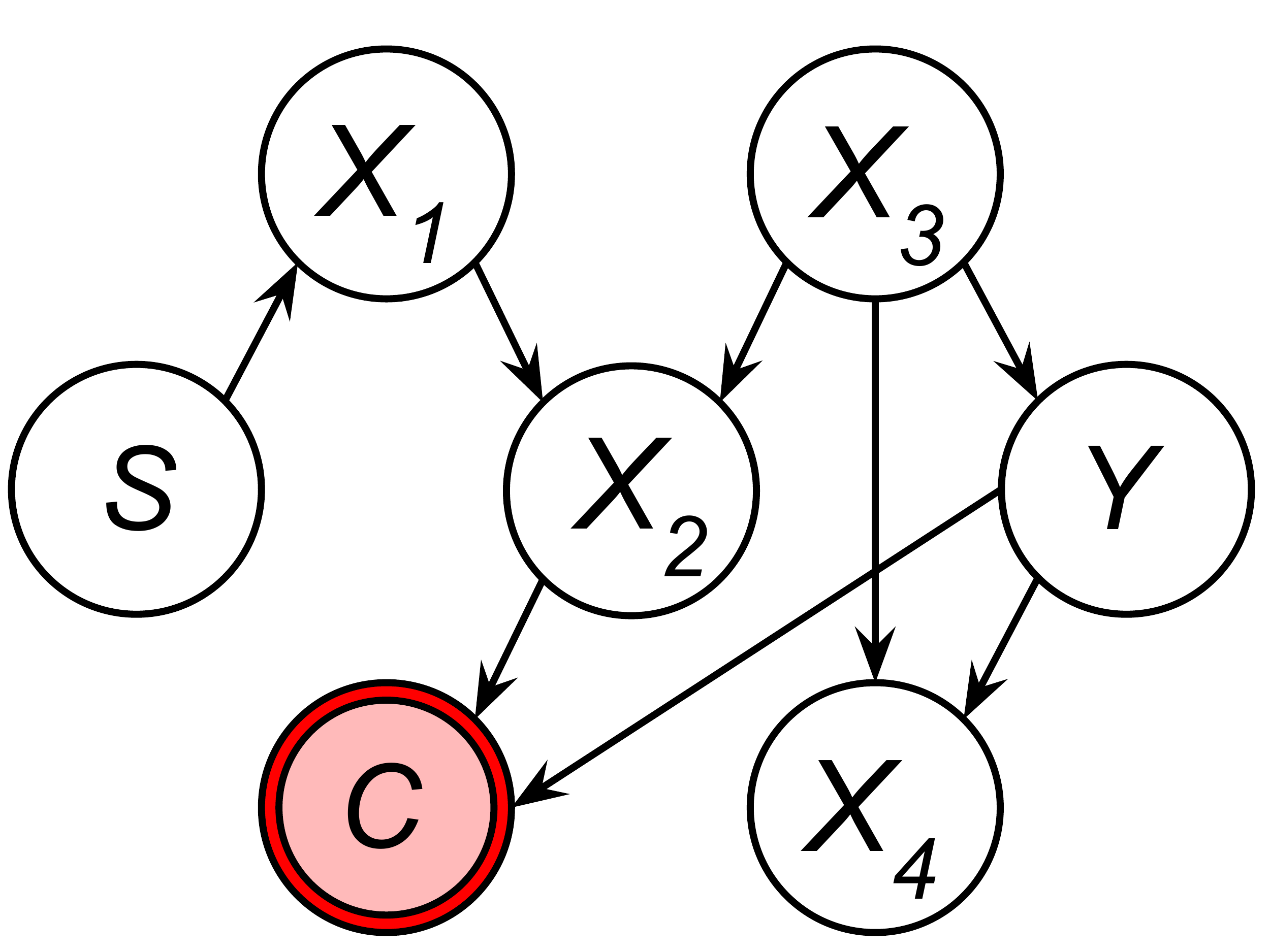}
         \caption{}
         \label{figure:structural_conditions:based_on_X2_Y}
     \end{subfigure}
     \hspace{.07cm}
     \begin{subfigure}[b]{0.23\linewidth}
         \centering
         \includegraphics[width=\textwidth]{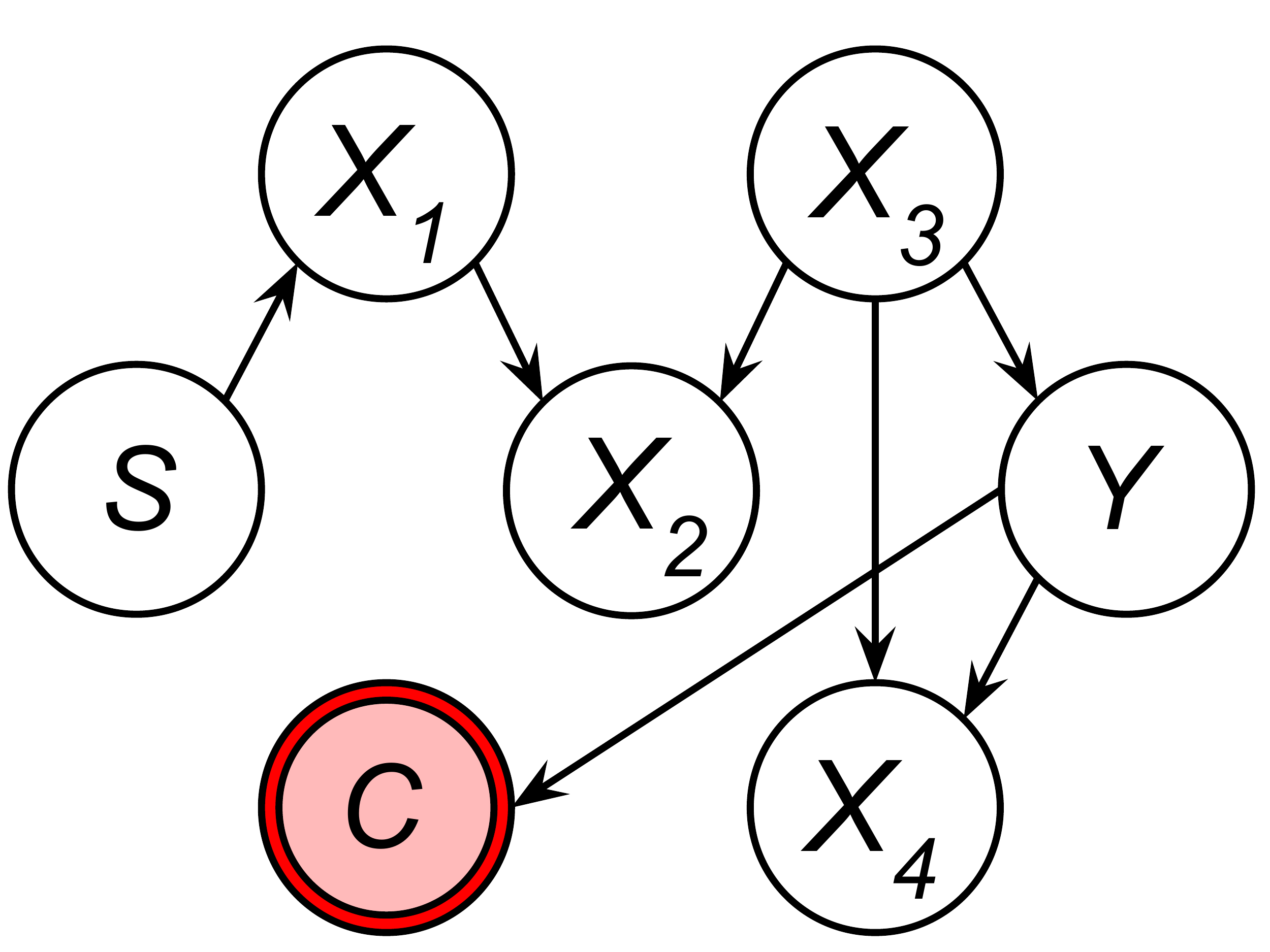}
         \caption{}
         \label{figure:structural_conditions:based_on_Y}
    \end{subfigure}
        \vspace{-0.8cm}
    \caption{\textmd{Examples that demonstrate our structural conditions.}}
    \vspace{-3mm}
    \label{figure:structural_conditions}
\end{figure}

In this section, we examine the relationship between selection bias and the fairness of predictive models. Specifically, we determine conditions under which selection bias may lead to unfair classifiers. To isolate the impact of selection bias from other sources of bias, such as bias due to finite data or bias due to ML model itself, we first define a fair data distribution that is free from any biases. 

\ifextended
By introducing selection bias in this context, we are able to clearly identify and study its effect on the fairness of the trained classifier, and establish sufficient and necessary conditions for when it may lead to discriminatory outcomes. 
\fi

\begin{defn}[{Fair data distribution}] \label{def:fairselectiondiagram} 
A data distribution $\dist$ that is compatible with a causal diagram $\cdag$ is considered {\em fair} for learning a classifier to predict outcome $\Yvar$ if any Bayes-consistent ML model trained on a sufficiently large sample from $\dist$ results in a fair classifier $\classifier$. Conversely, if this is not the case, the data distribution is considered {\em unfair} for learning $\Yvar$. 
\end{defn}

%




%
\ifextended
Our goal is to identify graphical conditions on the data collection diagram $\bcdag$ under which the trained classifier leads to unfair classification on the originally fair data distribution. Before we introduce our graphical conditions, we briefly discuss the notions of relevant features and Markov Boundary, which help identify features' roles in the decision-making process of a ML model.

Following~\cite{kohavi1997wrappers}, we say a feature $X \in \boldsymbol{X}$ is {\em strongly relevant} to a classification task if removal of $X$ from $\boldsymbol{X}$ leads to performance deterioration of the optimal Bayes classifier on $\boldsymbol{X} \in \Dom{\boldsymbol{X}}$. A feature X is {\em weakly relevant} if there exists a subset of features $\boldsymbol{Z} \subseteq \boldsymbol{X}$ such that the performance of the Bayes classifier on $\boldsymbol{Z}$ is worse than the performance on $\boldsymbol{Z} \cup \{X\}$. A feature is {\em irrelevant} if it is neither strongly nor weakly relevant. It can be shown that a variable $X$ is relevant for a classification task iff $X$ is in the {\em Markov Boundary} of $Y$, denoted $MB(Y)$, which consists of a set of variables $MB(Y) \subseteq \Allvar$ for which it holds that ($Y \indep \Allvar \mid MB(Y)$)~\cite{neapolitan2004learning}. 

Now we prove the following proposition that shows necessary and sufficient conditions for a fair data distribution.


%
\begin{proposition}\label{prop:unfair_process}
  A data distribution faithful to a causal diagram $\cdag$ is unfair for learning a classifier that predicts outcome $\Yvar$ iff
  there exists an attribute $X \in \features \setminus\Admis$ such that the following conditions C1 -- C2 hold.
\begin{itemize}
    \item [C1] There is an open path from $\ProtectedAttr$ to $X$ in $\cdag$ after conditioning on the admissible attributes $\Admis$, i.e.,  $X \not\indep_{\cdag} \ProtectedAttr \mid \Admis$.  
    \item [C2] The attribute $X$ is strongly relevant to $\Yvar$.
\end{itemize}
\end{proposition}
%

\ifextended
\begin{proof}
First we show that if $\classifier(\boldsymbol{x})$ violates \fairdef\ then
$\exists X \in {\boldsymbol{X}}$ such that conditions C1--C2 hold.
We consider the contrapositive: $\forall X \in {\boldsymbol{X}}$, either condition C1 or C2 do not hold, then we show that the classifier $\classifier$
is fair wrt. \fairdef
. More formally, if  
conditions $\neg C_1 \vee \neg C2$ hold, then $\pr_{\underlyingDist }(\classifier(\boldsymbol{x})=1\mid\privG, \admis)=\pr_{\underlyingDist}(\classifier(\boldsymbol{x})=1\mid\protG, \admis)$, $\forall {\boldsymbol{a}} \in \Dom{\boldsymbol{A}}$.

Let $\boldsymbol{X_1}$ denote the set of attributes that are independent of $S$ given the admissible variables in the underlying distribution, i.e., $X_i\indep_{G} S \mid \Admis$ for all $X_i\in \boldsymbol{X_1}$ (which is equivalent to the set of attributes satisfying $\neg$ C1) and $\boldsymbol{X_2}$ correspond to the attributes that are dependent on $\ProtectedAttr$ conditional on the admissible variables but the attribute is not relevant to the classifier, i.e., ($C1\wedge \neg$ C2). 

Since, $\neg (C1\wedge C2) \equiv (\neg C1)\vee (C1 \wedge \neg C2) $, we know that  all attributes  belong $X_i\in \boldsymbol{X_1} \cup \boldsymbol{X_2}$. The attributes in $\boldsymbol{X_2}$ are irrelevant to the classifier $h$ trained over the biased dataset. Therefore, the set of relevant attributes is a subset of $\boldsymbol{X_1}$, implying all relevant attributes (say $\bf{X_R} \subseteq \boldsymbol{X_1}$) are conditionally independent of $\ProtectedAttr$ in the causal graph $G$. Since $\underlyingDist$ is Markov compatible with $G$, $\boldsymbol{X_1} \indep_{\underlyingDist} \ProtectedAttr \mid \Admis$.
Therefore, 
$\PrPos_{\underlyingDist }(\classifier(\boldsymbol{x})\mid\privG, \admis)=\PrPos_{\underlyingDist}(\classifier(\boldsymbol{x})\mid\protG, \admis) =\PrPos_{\underlyingDist}(\classifier(\boldsymbol{x}) \mid \admis) $.

Now we show the converse, i.e, proving that if conditions C1 -- C2 hold, $\classifier(\boldsymbol{x})$ will violate \fairdef.  Let $\boldsymbol{X_1}$ denote the   features satisfying C1 -- C2, then $\boldsymbol{X_1} \subseteq {\boldsymbol{X_R}}$, the set of relevant attributes. Therefore, ${\boldsymbol{X}}_R \nindep_G S \mid {\boldsymbol{A}}$ implying that the classifier $h({\boldsymbol{x}})$ does not satisfy \fairdef (using faithfulness).
\end{proof}
\fi

Intuitively, the proposition states that if the protected attribute $\ProtectedAttr$ is not in the Markov Boundary of $\Yvar$ and is not correlated with any attributes $X$ in the Markov Boundary of $\Yvar$ after conditioning on $\Admis$, then collecting sufficiently large amounts of training data guarantees any Bayes-consistent ML model trained on it to be fair (see Example~\ref{exmaple-for-propositions} below).   
\fi
We now establish a graphical criterion on the data collection diagram $\bcdag$ such that training a classifier on the data suffering from selection bias leads to an unfair classifier. 

\vspace{-2mm}
\begin{proposition}\label{prop:unfair_selection_process}
A data distribution faithful to a data collection diagram $\bcdag$ is unfair for learning a classifier that predicts outcome $\Yvar$ iff either (1) the 
original data distribution compatible with $\cdag$ is unfair for learning a classifier, or (2) the following conditions C1 -- C2 hold.
\begin{itemize}
    \item [C1] The outcome variable $\Yvar\in \Pa(\SelectionVar)$.
    \item [C2] The selection variable $\SelectionVar$ is either a child of the protected attribute $\ProtectedAttr$ {\em or} there exists a variable $X \in \features \setminus\Admis$ such that the selection variable $X\in\Pa(\SelectionVar)$ and there is an open path between $X$ and $\ProtectedAttr$ that is not closed after conditioning on $\Admis$.
\end{itemize}
\label{Prop2:SelectionBias:structuralCondition}
\end{proposition}

\ifextended
\begin{proof}
We begin by proving that if case (1) or (2) holds, the classifier $\classifier$ would violate \fairdef.
If case (1) holds, the data distribution compatible with $\cdag$ is originally unfair. Therefore,we will have: $\exists X_i\in\Xvar\setminus\Admis$ such that $X_i\not\indep_{G}\ProtectedAttr\mid \Admis$ and $X_i\in MB(\Yvar)$ in $\cdag$ (\Cref{prop:unfair_process} ). As introducing a collider $\SelectionVar$ will not remove any variable out of $MB(\Yvar)$, we have: $X_i\in MB(\Yvar)$ in $\bcdag$, which means $X_i$ will be used by $h$ for prediction. Therefore, $\classifier(\boldsymbol{x}) \ \not\indep_{\cdag} \ProtectedAttr \mid \Admis$, the classifier $\classifier$ is unfair on the test data collected based on the causal diagram $\cdag$.
If case (2) holds, based on the definition of Markov boundary,  $\exists X_i \in MB(Y)$ in $\bcdag$, and for this $X_i$ we have $X_i \not\indep \ProtectedAttr \mid \Admis$, then $\classifier$ will be unfair according to \fairdef.

We now prove that if $\classifier$ violates \fairdef, then either the original data distribution, which is compatible with $\cdag$, is unfair (case (1)), or the conditions C1--C2 must hold (case (2)). Since the classifier $\classifier$ is Bayes-risk consistent, it is guaranteed to converge to a Bayes optimal classifier in the presence of sufficient amounts of data. All attributes that are used by $\classifier$, are in $MB(\Yvar)$ in the data collection diagram $\bcdag$ and all other attributes are not used by the learned classifier $\classifier$. 

Since $\classifier$ trained on the training data (compatible with $\bcdag$) violates \fairdef on the test data (compatible with $\cdag$), according to~\Cref{prop:unfair_process}, there must exist an attribute $X_i\in \Xvar\setminus\Admis$ (can be $\ProtectedAttr$), that is used by $\classifier$ and is dependent on $\ProtectedAttr$ after conditioning on $\Admis$ in $\cdag$.
More formally, $\exists X_i\in MB(\Yvar)$ in the graph $\cdag$, such that $X_i\not \indep_{\cdag} \ProtectedAttr \mid \Admis$.


As introducing a collider $\SelectionVar$ can expand $MB(\Yvar)$ while having $\Yvar$ as a parent, there are two possible cases that exist in the original data collection diagram $\cdag$: $X_i\in MB(\Yvar)$ or $X_i\notin MB(\Yvar)$.
If $X_i\in MB(\Yvar)$ in $\cdag$, $h$ will violate \fairdef before the introducing $\SelectionVar$, and this corresponds to the satisfaction of case (a) according to~\Cref{prop:unfair_process}.
And if $X_i\notin MB(\Yvar)$, i.e. $X_i$ is added to $MB(\Yvar)$ after introducing $\SelectionVar$ in $\bcdag$, $X_i$ cannot be a direct child or parent of $\Yvar$ because $C$ is a collider. That means $X_i$ can only be the parent of some child of $\Yvar$ in $\bcdag$ (condition C1). And since $X_i\notin MB(\Yvar)$ in $\cdag$, $X_i$ will be added to $MB(\Yvar)$ only if it is a parent of new child of $\Yvar$, which is $C$.
Since $X_i$ has the property: $X_i\not \indep_{\cdag} \ProtectedAttr \mid \Admis$, condition C2 is also satisfied. Therefore we proved the classifier's violation of \fairdef will correspond to either case (1) or (2).
\end{proof}
\fi

\ifextended
\begin{example} \label{exmaple-for-propositions}
We illustrate Proposition~\ref{prop:unfair_selection_process} with the data collection diagrams in \Cref{figure:structural_conditions}. The underlying causal diagram $\cdag$ that generates the test dataset is the same for all four sub-figures. The selection node $\SelectionVar$ represents the selection procedure and is decided based on its parents. Each figure represents a different selection mechanism. We can see that since $MB(Y) = \{X_3, X_4\}$, and since a Bayes optimal classifier $\classifier$ makes predictions based only on $MB(\Yvar)$, and there is no open path from $MB(\Yvar)$ to $\ProtectedAttr$, the original distribution compatible with $\cdag$ is a fair distribution. The training dataset will be a sample of the data collection process that is affected by selection bias, i.e. sub-figures in Figure~\ref{figure:structural_conditions}.

When the set of admissible variables is empty ($\Admis=\emptyset$),
\Cref{figure:structural_conditions:based_on_X2_Y} is the only case that would introduce unfairness; this corresponds to the condition explained in \Cref{prop:unfair_selection_process}, which we proved would introduce unfairness to an otherwise fair distribution. Specifically, as the original data distribution compatible with $\cdag$ is fair, case (1) in \Cref{prop:unfair_selection_process} is not satisfied for all graphs. As for conditions in case (2), only \Cref{figure:structural_conditions:based_on_X2_Y} and \Cref{figure:structural_conditions:based_on_Y} satisfy C1. Between these two structures, in \Cref{figure:structural_conditions:based_on_Y}, $\SelectionVar$ has only 1 parent; thus, it is impossible to satisfy C2,  which requires another parent $X\in\Xvar\setminus\Admis$. In \Cref{figure:structural_conditions:based_on_X2_Y}, we can let $X=X_2$, which becomes the child of $\Yvar$ after selection (C1) and is dependent on $\ProtectedAttr$ in the original data distribution conditioning on $\Admis=\emptyset$ (C2). 
In the experiments, we will empirically show that~\Cref{figure:structural_conditions:based_on_X2_Y} is the only selection procedure that would introduce unfairness to this data collection process that originally generates fair data distributions.
\end{example}
\fi

\ifextended
\Cref{prop:unfair_selection_process} requires that the data distributions is faithful to the data collection diagram $\bcdag$ for proving that the conditions (1) or (2) imply unfairness of the classifier. If faithfulness is violated, unfairness of the classifier would still imply that either (1) or (2) holds.
\fi

An important conclusion from this section is that when discrimination arises only from selection bias, the selection of data points is influenced by the training label $Y$. This leads to the violation of the independence assumption $Y \not \indep C \dsep \mb X$. However, most techniques in ML for addressing selection bias assume that selection bias leads to covariate shift, i.e., $Y  \indep C \dsep \mb X$. Hence, these techniques cannot handle fairness issues stemming from selection bias. In contrast, \sys can handle situations where the selection bias is label-dependent, offering solutions not provided by other methods.

%% file: Section_Tex_Files/experiments.tex
\begin{table*}[t] \centering \small
{\footnotesize
	\label{tbl:data}
		\begin{tabular}{@{}lrrrrrrrrrrrrrrrrrr@{}}\toprule
			{} & {Dataset} & {Att. [$\#$]} & {Rows[$\#$]} & \textsc{Orig} & \lfr & \reweighing & \advdebias & \covofdist & \ipw & \minmax & \sysBound &  \sysComplete & \sysMissA & \sysMissPartU  & \\ \midrule
			{} & \textbf{Adult} & 8 & 45k & 0.8s & 56.5s & 4.6s & 5.6s & 1.1s {} & 4.6s & 8m5s & 1.8s & 2.5s & 2.7s & 5.3s &  \\ 
			{} & \textbf{Law}& 11 & 18k & 0.6s & 43.3s & 11s & 3.7s & 0.8s {} & 10.9s & 3m17s & 3.4s & 5.7s & 6.1s & 12s & \\ 
		    {} & \textbf{HMDA}& 8 & 3.2m & 6.7s & 45m32s & 9m & 1m57s & 8.9s & 8m57s & \multicolumn{1}{c}{DNF} & 16.8s & 4m5s & 4m8s & 8m45s & \\ 
		    {} & \textbf{Syn}& 5 & 200k & 1.8s & 4m36s & 56.5s & 23.2s & 3.1s & 56.5s & 37m24s & 15.6s & 28.5s & 28.7s & 59.4s &\\
		    \bottomrule
		\end{tabular}
  \vspace{-2mm}
		\caption{\textmd{Average runtime in seconds for $\sys$ and baselines. Results of \minmax on HMDA are omitted due to the huge runtime of the used KDE method while applying on large data (did not finish in 100hrs, expected time required >1 year).}
		\vspace{-10mm}
        \label{tab:datasets}}
	}
\end{table*}

\begin{figure*}
    \centering
    \includegraphics[scale=0.33]{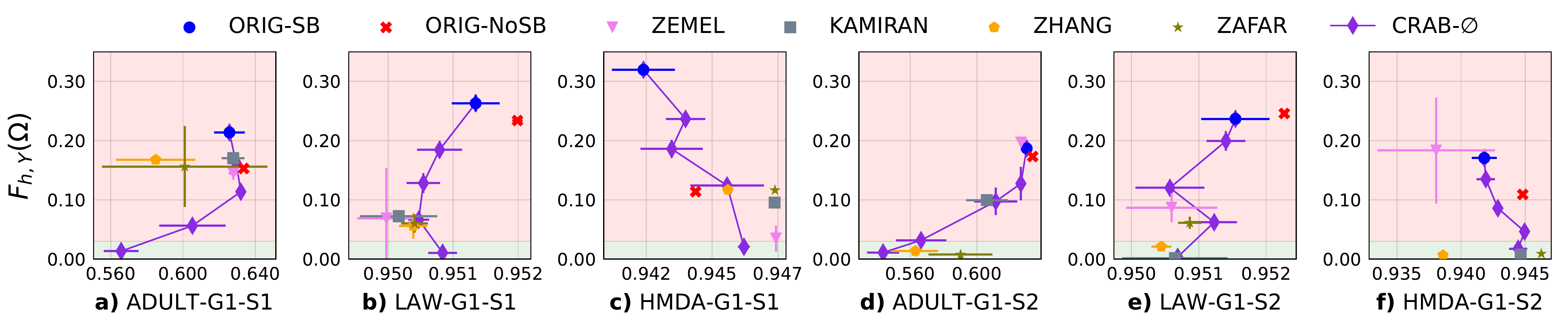}
    \vspace{-4mm}
    \caption{\textmd{Equality of opportunity (y-axis)-F1-Score (x-axis) comparison for \sysBound and baseline methods in the absence of external data. (a to c) correspond to scenario 1, where $\CondDemoParityb{\Yvar}$ largely deviates from $\CondDemoParity{\Yvar}$; (d to f) correspond to scenario 2, where $\CondDemoParityb{\Yvar}\approx\CondDemoParity{\Yvar}$. Selection variable $C$ is placed as a child of $S$ and another attribute $X\in\Allvar$ (same as Figure~\ref{figure:structural_conditions:based_on_S_X4}).} }
    \label{fig:exp:abs_ext_info_comparison} 
    \vspace{-3mm}
\end{figure*}
\label{sec:experiments}
We assess how selection bias impacts classifier fairness and demonstrate the efficacy of \sys{} in achieving fairness while preserving accurate predictions. The code and input files used to generate the results can be found at~\cite{crab_code}. We aim to address the following questions. \textbf{Q1}: Can \sys{} leverage 
    varied amounts of external data to
    guarantee fairness in the presence of selection bias?  How does it compare to the state-of-the-art fair classification methods and current techniques in handling  selection bias? (Section~\ref{sec:exp:comparison}). \textbf{Q2}: When does selection bias introduce unfairness in scenarios where the unbiased data generative process is fair?  
    \textbf{Q3}:  How does \sys{} adapt to other fairness metrics and classification techniques? (Section~\ref{sec:exp:sensitivity})

\vspace{-0.6cm}
\subsection{Setup}

\noindent\textbf{Datasets.}
We used the following datasets: \textbf{Adult} contains financial and demographic data to predict if an individual's income exceeds \$50K, with gender as the protected attribute. \textbf{Law} contains law school student data to predict exam outcome (Pass/Fail), with race as the protected attribute. \textbf{HMDA} contains mortgage application data to predict loan approval or denial, with race as the protected attribute. \textbf{Syn} is synthetic data generated based on~\Cref{figure:structural_conditions}.



We introduced selection bias in our datasets by simulating six different mechanisms, by adding the selection variable as a child of various attribute sets and varying the selection probability. This bias was applied to both the training and validation sets, while the test data remained untouched. We repeated the experiments five times and report the average and standard deviation for each method. Our evaluation of the techniques includes statistical parity and equality of opportunity as the fairness measures. A lower value of $\sCondDemoParity$ and a higher F1-Score indicate better performance.

\noindent \textbf{ML Models.} We evaluate with logistic regression (LR), Support Vector Machine (SVM) with a linear kernel, and neural network (NN) models, implemented using PyTorch~\cite{paszke2019pytorch}. Logistic regression was used as the default classifier unless otherwise specified.


\noindent \textbf{External Data Sources.} We evaluated \sys\ under the following settings of access to external data. (1) \sysBound, with no external data to use Proposition~\ref{proposition:no_external_data:dp:max_min_bound}; (2) \sysComplete, with sufficient external data to apply Proposition~\ref{prop:exactcal}; (3) \sysMissPartU, with limited external data to utilize Proposition~\ref{proposition:limited_external_data:dp:max_min_bound}; and (4) \sysMissA{}, with no external data about admissible attributes to apply Proposition~\ref{proposition:limited_external_data:dp:exactcalc}.

\noindent\textbf{Baselines.}
We compare \sys\ with the following representative baselines: \lfr~\cite{zemel2013learning} and \reweighing~\cite{kamiran2012data} are pre-processing methods that modify the training data to obscure information about the protected attributes. \advdebias~\cite{zhang2018mitigating} and \covofdist~\cite{zafar2017fairnessbeyond,zafar2017fairnessconstraints} are in-processing methods that maximize model quality while minimizing fairness violation. \ipw is an inverse propensity score weighting (IPW) based pre-processing method that utilizes external data to recover from selection bias~\cite{cortes2008sample}.
\minmax\ is an in-processing method that minimizes the worst-case unfairness with respect to equality of opportunity for Logistic Regression~\cite{rezaei2020robust}.
\ipw and \minmax are techniques used to address dataset shift from domain adaptation, and are applied only in the presence of external data.
We utilized the IBM's AI Fairness 360~\cite{bellamy2018ai} implementation for the pre-processing methods and the original implementation of \minmax\cite{rezaei2020robustgithub}. Other baseline methods were implemented from scratch in our framework to ensure a fair comparison.
\vspace{-0.3cm}
\subsection{Solution Quality and Fairness Comparison}\label{sec:exp:comparison}
The performance of \sys{} and the baselines was evaluated for varying levels of access to external data. In all figures, the region with a green background indicates that the fairness requirements were satisfied, while the red region indicates an unfair region.
\subsubsection{Absence of external data}\label{sec:exp:abs-info}
We compared \sysBound with the fair ML baselines. We varied the fairness requirement $\tau$ from 0 to $\theta$, where $\theta$ was the unfairness of the classifier trained on the biased dataset. To simulate two different scenarios, we introduced the selection node $\SelectionVar$ as a child of $\ProtectedAttr$ and another variable $X_2$, and varied the selection criterion, $\pr_{\underlyingDist}(\SelectionVar=1\mid\Pa(C))$. In the first scenario (S1), the biased dataset had a much lower level of unfairness compared to the test data, making it necessary to consider the selection bias to ensure fairness over the test data. In contrast, the biased dataset was approximately as fair as the unbiased test data in the second scenario (S2), making it possible to enforce fairness on the biased training data to ensure fairness on the unbiased test data.

\begin{figure*}
    \centering
    \includegraphics[scale=0.3]{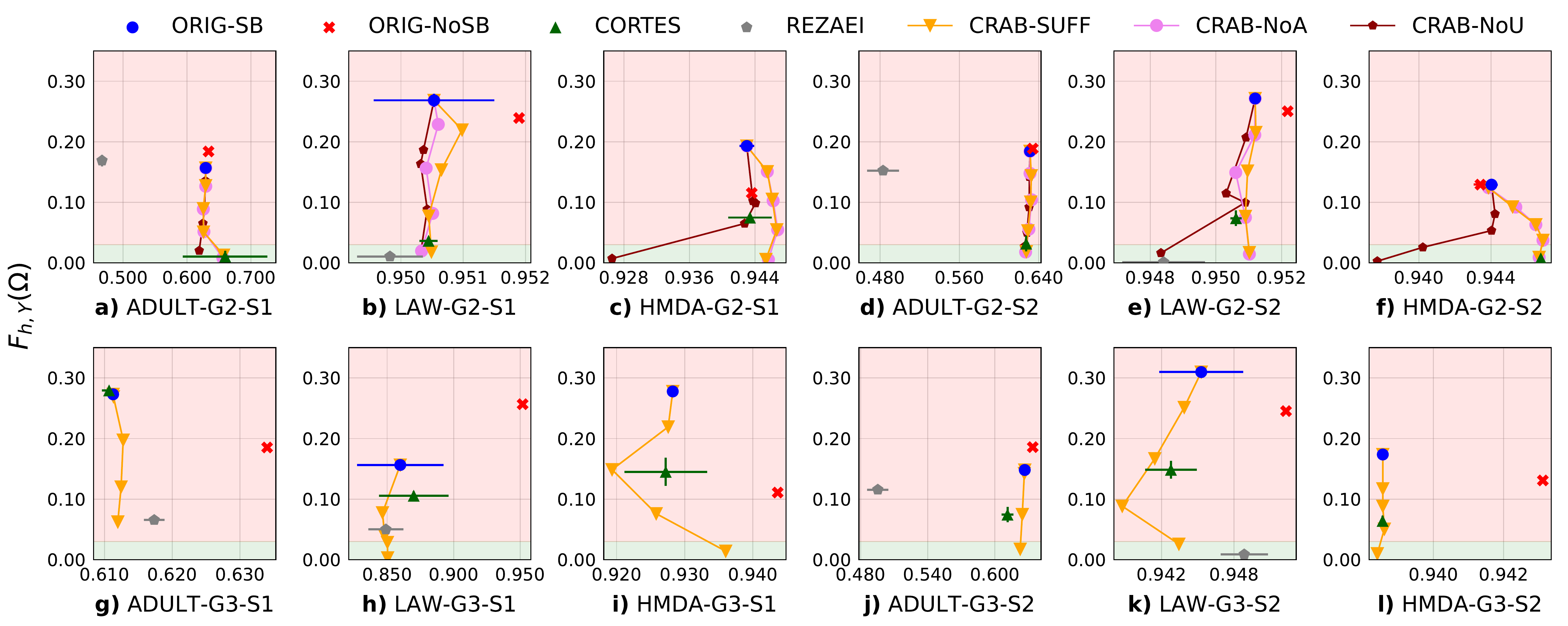}
    \vspace{-4mm}
    \caption{\textmd{Equality of Opportunity (y-axis)-F1-Score (x-axis) comparison for \sysComplete, \sysMissPartU, \sysMissA, \ipw, and \minmax given the existence of external data. (a to f) have a selection variable similar to \Cref{figure:structural_conditions:based_on_X2_X4}, and (g to l) have a selection variable similar to \Cref{figure:structural_conditions:based_on_X2_Y}. S1 corresponds to the scenario where $\CondDemoParityb{\Yvar}$ largely deviates from $\CondDemoParity{\Yvar}$, and S2 denotes the scenario where $\CondDemoParityb{\Yvar}\approx\CondDemoParity{\Yvar}$. Error Bars for \sysComplete, \sysMissA and \sysMissPartU are omitted for clarity. Results of \minmax on HMDA are omitted due to huge runtime.}}
    \vspace{-5mm}\label{fig:exp:with_ext_info_comparison}
\end{figure*}
\noindent \textbf{Solution quality.} Figure~\ref{fig:exp:abs_ext_info_comparison} compares the fairness of the trained classifier and F1-Score. We observe that \sysBound learns a fair classifier ($\CondDemoParity{\Yvar}<0.05$ for $\tau=0$ corresponding to the point with lowest y-coordinate) for all datasets across both scenarios. In fact, \sysBound{} achieves perfect fairness while incurring a very low loss in F1-Score for the Law and HMDA datasets.
In contrast, most of the baseline methods fail to completely remove the model's discrimination in most of the cases.  
In certain cases, baseline techniques like \lfr  improve fairness (HMDA-G1-S1), but the same technique returns a highly unfair classifier for HMDA-G1-S2.  This is because the selection mechanism creates a false sense of fairness on training data ($\CondDemoParityb{\Yvar} \approx 0$),   while $\CondDemoParity{\Yvar}$ remains high.
This is the reason that most of the baselines achieve worse fairness for the first scenario compared to the second one. It shows that the two scenarios behave differently even though the selection bias is a function of the same set of attributes (causal diagram does not change), with the only difference being the selection probabilities. However, \sysBound{} upper bounds $\CondDemoParity{Y}$, which helps to enforce fairness across both settings for all datasets. 
\begin{mdframed}
\textit{Key Takeaway.} \sysBound{} (with fairness requirement threshold $\tau=0$) achieves perfect fairness for all scenarios and datasets, while baseline methods demonstrate aberrant behavior. 
\end{mdframed}

\noindent \textbf{Comparison between \origtest and \sysBound{}.} Figure~\ref{fig:exp:abs_ext_info_comparison} demonstrates that \sysBound{} outperforms \origtest in fairness whenever it achieves an F1-score comparable to that of \origtest (red point has higher equality of opportunity compared to \sysBound{} point at same x-coordinate). The only case where \origtest has marginally higher accuracy 
than \sysBound{} is Law-G1-S2 with an insignificant accuracy difference ($<1\%$). This highlights that \sysBound{} is capable of achieving an F1-score comparable to that of \origtest even without external data to address selection bias.

\revc{\noindent\textbf{Quality vs. fairness tradeoff.} Unlike most prior fair ML techniques, \sys{} allows the user to specify a fairness requirement $\tau$, which allows to simulate varying needs. Figure~\ref{fig:exp:abs_ext_info_comparison} shows that \sysBound{} achieves the same fairness and accuracy as \orig when $\tau$ exceeds the fairness bound in Theorem~\ref{proposition:no_external_data:dp:max_min_bound}. On reducing $\tau$, \sysBound{}'s fairness improves consistently with a minor or no loss in F1-score until $\tau > 0.1$. Further reducing $\tau$ to $0$ worsens the F1-score for the Adult dataset by around $8\%$ but less than $1\%$ for all other datasets. 
We observe that \sysBound{} and baseline techniques achieve similar F1-scores when \sysBound{} is configured to achieve similar fairness.
This demonstrates \sysBound{} ability to match baseline performance while achieving perfect  fairness by achieving zero $\CondDemoParity{Y}$.

\begin{mdframed}
\textit{Key Takeaway.} \sysBound{} considerably improves fairness of the trained classifier, with only a minor loss in F1-Score.
\end{mdframed}}

\vspace{-0.1cm}
\subsubsection{Availability of external data}\label{sec:exp:ext-info}
In this experiment, we assess the performance of \sys{} under different levels of availability of external data. We consider three different cases. (1) \textbf{Sufficient information:} this setting is applicable when external data source about some statistics of the unbiased distribution are available. \sysComplete{} uses estimates of $\mbox{$\pr_{\underlyingDist}(\boldsymbol{u}\mid s,\boldsymbol{a})$}$ for model training. (2) \textbf{Missing ${\boldsymbol{A}}$}: this setting is applicable when external data source about ${\boldsymbol{A}}$ is not available. For example, when ${\boldsymbol{A}}$ is the prediction target ${\boldsymbol{A}}=\{Y\}$, we cannot access labels for the unbiased distribution. (3) \textbf{Missing ${\boldsymbol{U}}$}: this setting considers availability of partial external data source about a set of attributes ${\boldsymbol{U}}$. For a fair comparison, we compare \sys{} under these settings with baselines that use external data (\ipw and \minmax). 
\revb{Note that \ipw requires additional information compared to \sysComplete{}, which relies only on the estimates of $\pr_{\underlyingDist}(\boldsymbol{u}\mid s,\boldsymbol{a})$ computed from unbiased external data. In contrast, \minmax requires unlabeled external data, which is similar to the setting of \sysMissA{}.}
\begin{figure}
    \centering
    \includegraphics[scale=0.24]{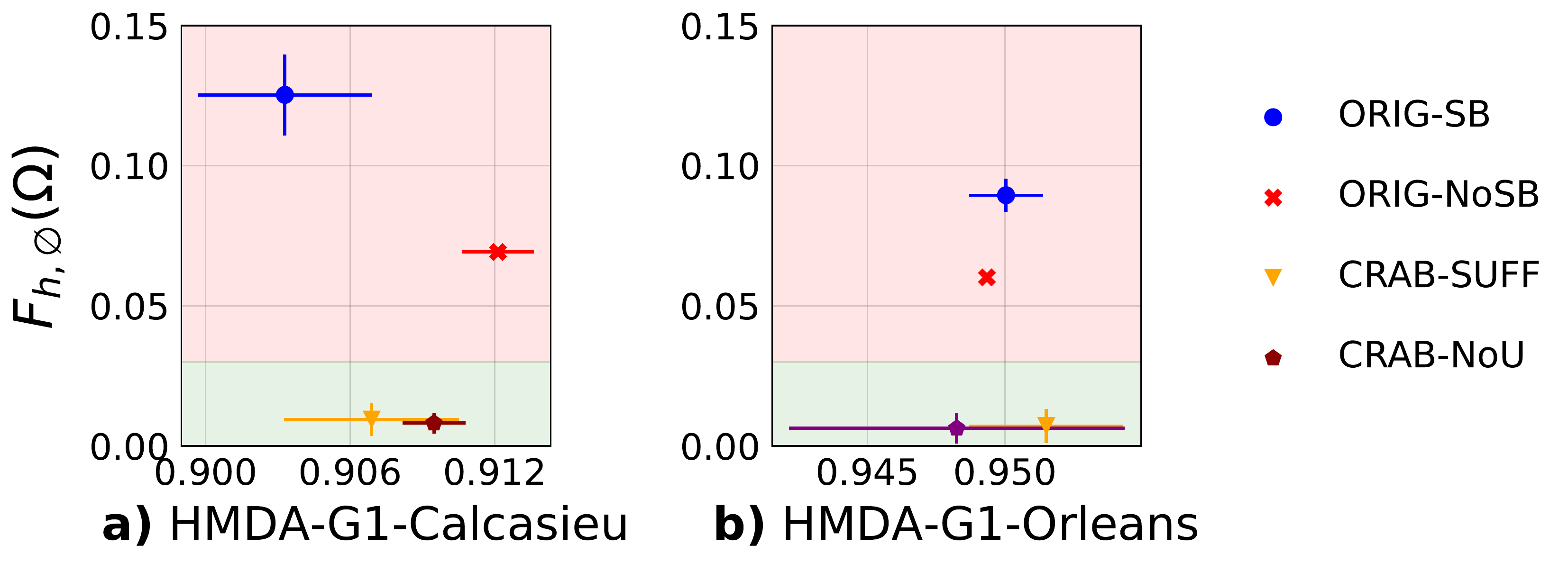}
    \vspace{-4mm}
    \caption{\textmd{Statistical Parity (y-axis)-F1-Score plots (x-axis) for two parishes in Louisiana where \sysComplete and \sysMissPartU use the real-world census to provide external data for training.}}
    \vspace{-5mm}
    \label{fig:exp:real_ratio_comparison}
\end{figure}



\noindent \textbf{Case Study: Louisiana.} We examine the effects of selection bias in the HMDA dataset for two parishes in Louisiana: Calcasieu and Orleans. Using public census data~\cite{statisticalatlas} as the source of external data, we ran \sysComplete{}, \sysMissA{}, and \sysMissPartU{} with $\tau$ set to $0$. Selection bias was introduced based on age and race.  Despite the real-world ratios not being exactly consistent, \sys{} successfully trains a fair model (Figure~\ref{fig:exp:real_ratio_comparison}). Furthermore, the F1-Score of the trained model is higher than that of \orig in Calcasieu. This experiment demonstrates the potential of \sys{} to use public data to train fair ML models. Next, we compare the performance of \sys{} with other baselines and varied settings of external data.

\begin{figure}
    \centering
    \hspace{-2mm}
    \includegraphics[scale=0.24]{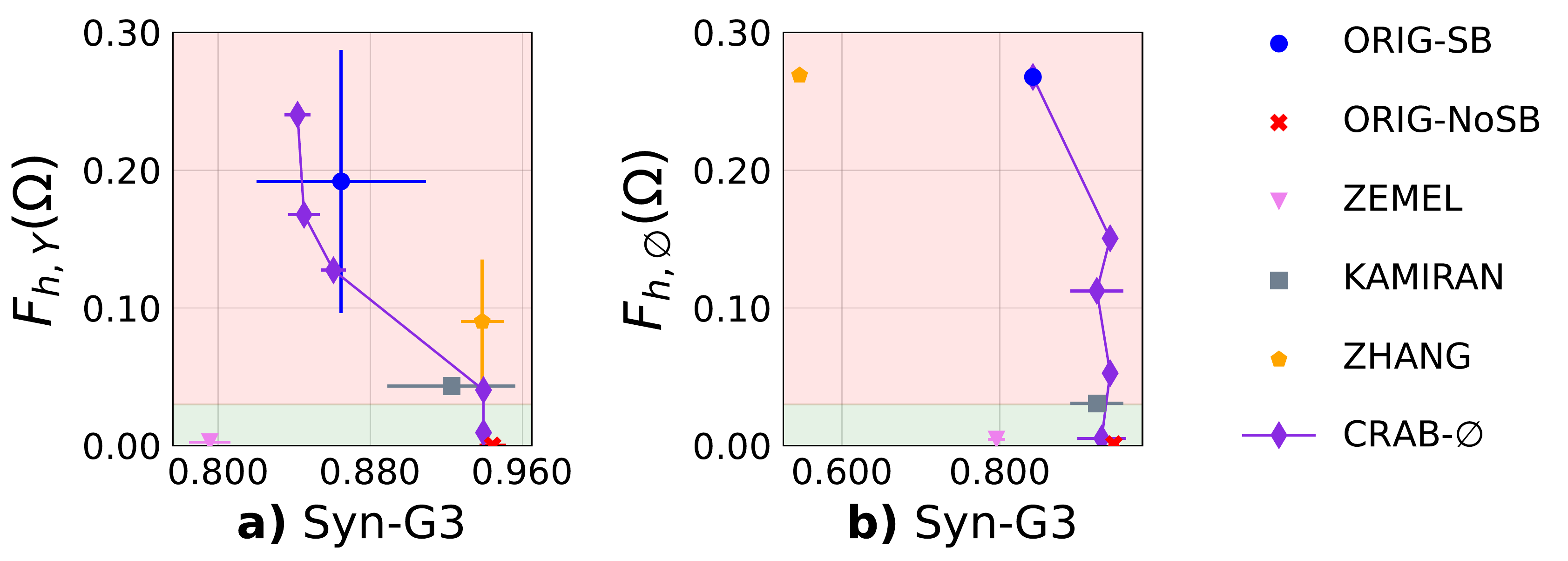}
    \vspace{-0.1cm}
    \caption{\textmd{A Discrimination (y-axis)-F1-Score (x-axis) comparison for \sysBound and baseline methods evaluated on Syn. \covofdist was omitted due to its extremely low F1-Score in this experiment.}}
    \vspace{-5mm}
    \label{fig:exp:syn_comparison}
\end{figure}

\noindent \textbf{Solution quality.} The comparison between \sys{} and other baselines, including \ipw, \minmax, and \origtest, is shown in Figure~\ref{fig:exp:with_ext_info_comparison}. The results demonstrate that all \sys{} methods attain approximately zero equality of opportunity for $\tau=0$, with only minimal reductions in F1-Score. In fact, the quality of the fairest classifier is higher than the fairness-agnostic classifier trained on the original data (HMDA-G2-S2 and HMDA-G3-S1). We tested the fairness-accuracy tradeoff in detail in Figure~\ref{fig:exp:syn_comparison} and show that ensuring fairness can indeed improve overall classifier performance.
In contrast, \ipw{} and \minmax baselines use additional external data to recover from selection bias but remain unstable with respect to fairness.  \ipw{} helps to ensure fairness in Figure~\ref{fig:exp:with_ext_info_comparison} (a to d), but the trained classifier is highly unfair in Figure~\ref{fig:exp:with_ext_info_comparison} (g to l). \ipw{} relies on the estimation of propensity scores, which are highly dependent on the quality of classifiers learned to estimate $\pr_{\underlyingDist}(\SelectionVar=1\mid\Xvar=x)$ and $\pr_{\underlyingDist}(\SelectionVar=1\mid Y=1,\Xvar=x)$. Noisy estimation of these probabilities affects \ipw{} performance. {Similarly, \minmax fails to ensure fairness on many cases due to inaccurate density ratio estimation. Note that both \sysMissA{} and \minmax require unlabeled external data, while \sysMissA{} outperforms \minmax in terms of both accuracy and stability (Figure~\ref{fig:exp:with_ext_info_comparison} (a to f)). }

\sysMissPartU{} is a bound-based approach, meaning a loose upper bound for $\CondDemoParity{Y}$ could result in excessive fairness restrictions. The superiority of bound-based or estimation-based approaches cannot be determined universally. However, if the difference between the upper bound and $\CondDemoParity{Y}$ is significant, using a bound-based approach like \sysMissPartU{} may negatively impact classifier performance to ensure fairness, as shown in Figure~\ref{fig:exp:with_ext_info_comparison} (f).


\noindent \textbf{Effect of varying external data}. Out of three \sys methods, \sysMissPartU requires the least external data, followed by \sysMissA and \sysComplete. However, these two methods are not shown in Figure~\ref{fig:exp:with_ext_info_comparison} (g-l) (Graph G3) as they are not suitable for this scenario.
In Figures~\ref{fig:exp:with_ext_info_comparison} (a-f), both \sysComplete{} and \sysMissA{} achieve slightly higher F1-Scores for the same level of fairness compared to \sysMissPartU{}.
\sysMissPartU{} eliminates bias by setting a cap on $\CondDemoParity{Y}$ instead of estimating it, which may lead to a non-monotonic skyline as seen in Law-G2-S2. Nonetheless, \sys{} ensures fairness across all scenarios regardless of the level of access to external data.

\vspace{1mm}
\begin{mdframed}
\textit{Key Takeaway.} \sysComplete{} produces the most accurate model with zero equality of opportunity. \revb{Despite \ipw{} utilizing more external data and \minmax having the same level of external data access as \sysMissA{}, these methods do not consistently result in a fair classifier.}
\end{mdframed}

\noindent \textbf{Running time.} The running time comparison between \sys, \textsc{Orig} and all baselines with the logistic regression classifier is presented in Table~\ref{tab:datasets}. The results show that \sys{} takes less than 10 seconds for small datasets (Adult and Law), and under 5 minutes for larger datasets like HMDA (except \sysMissPartU which is slightly slower). Among the baselines, \minmax had the longest running time, taking over 24hrs for the HMDA dataset. Although faster than \minmax, \lfr still took about $10\times$ longer than \sys. The other baselines had a similar execution time as \sys{}.  



\ifextended

\subsection{Effect of Selection Bias on Fairness and F1-Score}
In this section, we conduct empirical evaluations to support the findings discussed in Section~\ref{section:introducing_bias_out_of_the_blue} and demonstrate how selection bias in data collection can lead to unfair outcomes. Additionally, we assess the impact of enforcing fairness on the model's performance.

\noindent \textbf{Unfairness due to selection bias.}\label{sec:experiments:out_of_blue}
In this experiment, we consider the \texttt{Syn} dataset and introduce selection bias in five different ways (R denoting randomly and G1 to G4 corresponding to the procedures in Figure~\ref{figure:structural_conditions}(a to d)) to evaluate its impact on fairness. The test distribution of the \texttt{Syn} dataset satisfies $Y\indep S$ ($\CondDemoParity{Y}=0.001$), but the independence does not hold in the biased training dataset for scenarios G1 to G3, where $Y\nindep S \mid C$. We trained the logistic regression classifier on the biased dataset and evaluated its accuracy and unfairness on the unbiased test set (Table~\ref{table:results_unfairness_out_of_blue}). The only case where selection bias causes a notable increase in unfairness is when the selection node is a child of $Y$ and another variable $X$ such that $S\not\indep X\mid Y$ ($\CondDemoParity{Y}$ corresponds to $\Admis=\{Y\}$). We observe similar results for the SVM classifier.
This empirical observation conforms with our analysis in \Cref{Prop2:SelectionBias:structuralCondition}.

\begin{table}[t]
\small
\centering
 \begin{tabular}{|c | c|c | c | c| c |} 
 \hline
 Scenario  & $Y\in \texttt{Pa}(C) $&$\texttt{Pa}(C)  \nindep S$ & F1-Score & $\CondDemoParity{\Yvar}$ \\ [0.5ex] 
 \hline\hline
  R & No & No & 0.957 & 0.001\\\hline
 G1 & No&Yes& 0.957 & 0.017\\\hline
G2 & No&Yes& 0.957 & 0.015 \\\hline
G3 & Yes&Yes& 0.850 & \textbf{0.235} \\\hline
 G4 & Yes&No& 0.957 & 0.002\\\hline
 \end{tabular}
 \caption{\textmd{Comparison of F1-Score and unfairness when varying the selection bias procedure. R denotes a random procedure, and G1 to G4 correspond to the graphs in Figure~\ref{figure:structural_conditions}.}\label{table:results_unfairness_out_of_blue}}
 \vspace{-5mm}
\end{table}
\noindent \textbf{Can fairness improve classifier quality?}  Figure~\ref{fig:exp:abs_ext_info_comparison} (c) and Figure~\ref{fig:exp:with_ext_info_comparison} (f) showed that the classifier trained by \sys{} can sometimes achieve higher F1-Scores than \orig and \origtest while ensuring fairness ($\CondDemoParity{Y} = 0$). To validate this observation, we evaluated \sys{} extensively on the Syn dataset where the unbiased dataset is fair ($\CondDemoParity{Y} = 0$). Figure~\ref{fig:exp:syn_comparison} shows that ensuring fairness can indeed improve classifier F1-Scores by more than $10\%$. However, most baselines remain either unfair or achieve much lower F1-Scores. This evaluation demonstrates that ensuring fairness can on occasion improve classifier quality, and \sys{} achieves the best F1-Score and fairness.
\vspace{1mm}
\begin{mdframed}
\textit{Key Takeaway.} Ensuring fairness can sometimes improve classifier quality.
\end{mdframed}
\fi
\subsection{Sensitivity to Parameters}\label{sec:exp:sensitivity}
In this section, we examine how \sys{} and other baselines respond to changes in fairness metric, ML model, and external data size.

\begin{figure}
    \centering
    \includegraphics[scale=0.24]{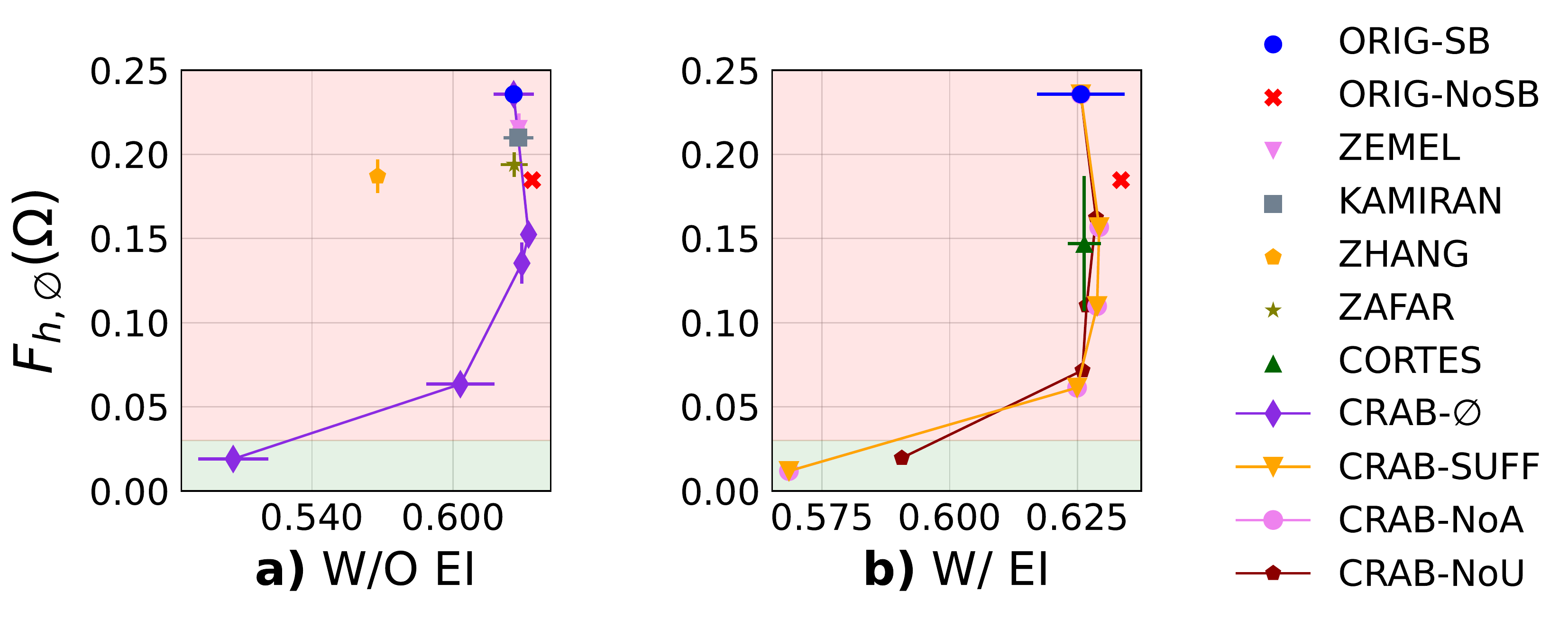}
    \caption{\textmd{Statistical Parity (y-axis)-F1-Score (x-axis) comparison for \sys and baseline methods for Adult-G1-S1.}} 
    \label{fig:exp:spd_comparison}
\end{figure}

\noindent \textbf{Fairness metric.}\label{sec:exp:spd}
Figure~\ref{fig:exp:spd_comparison} compares \sys{} on both settings, for the Adult dataset and for statistical parity. In figure~\ref{fig:exp:spd_comparison}(a), all baselines achieve statistical parity of $0.15$ except \sysBound{}, which achieves zero statistical parity. Further, we observe that \sysBound{} achieves the maximum F1-Score of $0.63$ while achieving a statistical parity of less than all other techniques, including the classifier trained on the unbiased original dataset. This demonstrates the effectiveness of \sysBound{} at achieving fairness even without any external data. Figure~\ref{fig:exp:spd_comparison} compares \sys{} under three different settings of access to external data (\sysMissA{}, \sysMissPartU{}, \sysComplete{}). We observe that \sys{} (with $\tau$ is set to zero) achieves statistical parity $\approx$ 0 across all settings. Further, the \ipw{} baseline does not achieve fairness even though it has access to unbiased data.
By comparing figures~\ref{fig:exp:spd_comparison}(a) and (b), it can be seen that when external data is used, \sysMissPartU{} is able to achieve an F1-Score of $0.59$ with zero statistical parity, as opposed to $0.52$ without external data. Additionally, both \sysMissA{} and \sysComplete{} reach an F1-Score above $0.57$. This demonstrates that incorporating any level of external data can improve the performance of the trained classifier.


\begin{figure}
    \centering
    \vspace{-5mm}
    \includegraphics[scale=0.24]{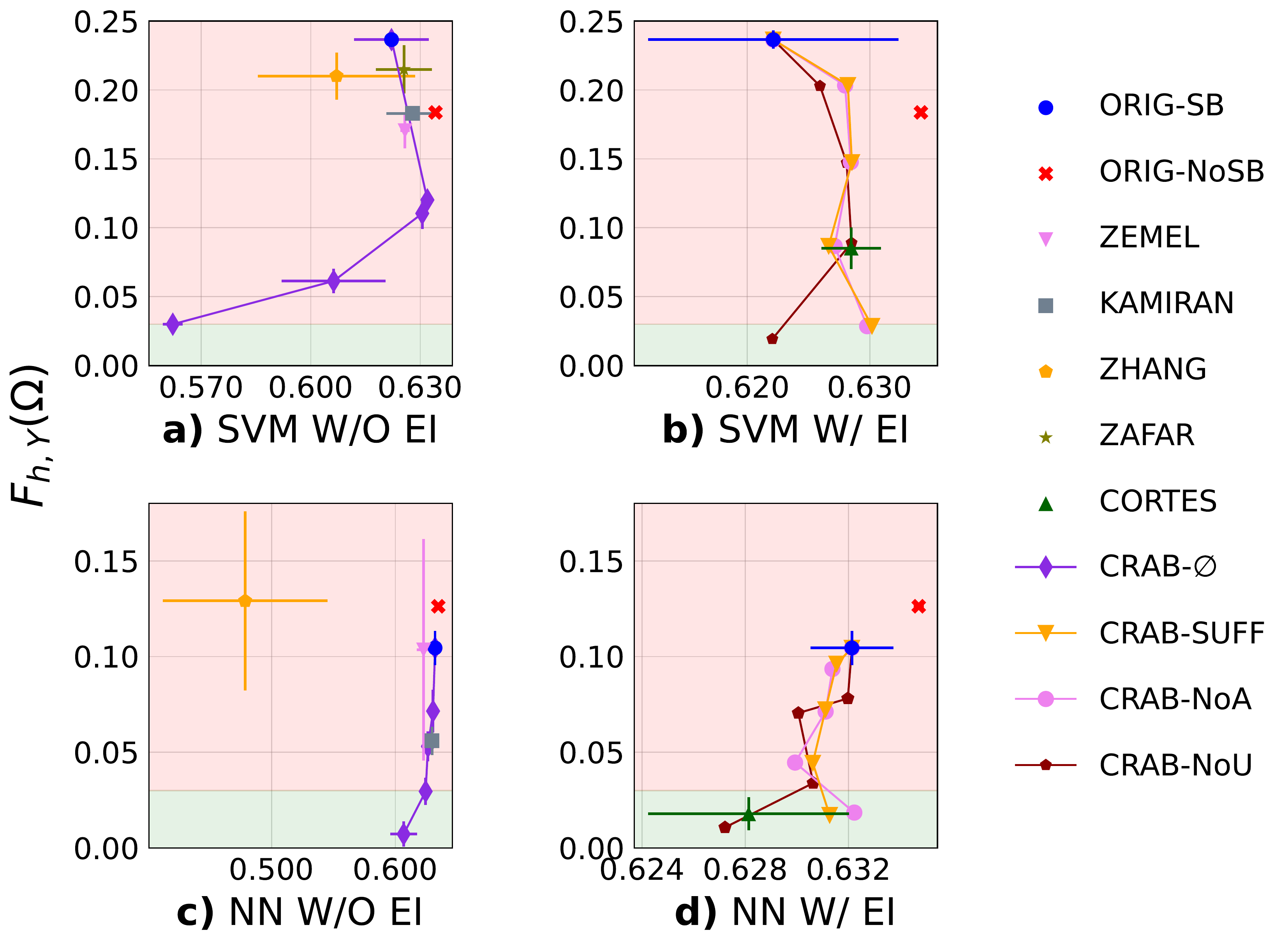}
    \vspace{-0.4cm}
    \caption{\textmd{Equality of Opportunity (y-axis)-F1-Score (x-axis) comparison for \sys and baseline methods on the Adult dataset with the selection mechanism described in \Cref{figure:structural_conditions:based_on_X2_X4}}.
    }
    \vspace{-0.4cm}
    \label{fig:exp:svm_nn_comparison}
\end{figure}
\noindent \textbf{ML Models.} \sys{} can be adapted to a variety of classification algorithms by modifying its loss function. Figure~\ref{fig:exp:svm_nn_comparison} shows that \sys{} produces fair results with both SVM and NN classifiers, while most of the baseline models still display unfairness ($\CondDemoParity{Y}>0.05$).
Further, \sysBound{} achieves the maximum F1-Score with the maximum fairness ($\CondDemoParity{Y}<0.12$ in Figure~\ref{fig:exp:svm_nn_comparison}(a)), while all baselines perform worse ($\CondDemoParity{Y}>0.17$). Comparing Figure ~\ref{fig:exp:svm_nn_comparison} (a) and (b), we observe that \sysComplete{} achieves higher F1-Scores than \sysBound{} while maintaining $\CondDemoParity{Y}<0.03$. We observe similar trends for the neural network (Figure~\ref{fig:exp:svm_nn_comparison} (c), (d)).

\begin{figure}
    \centering
    \hspace{-0.4cm}
    \includegraphics[scale=0.15]{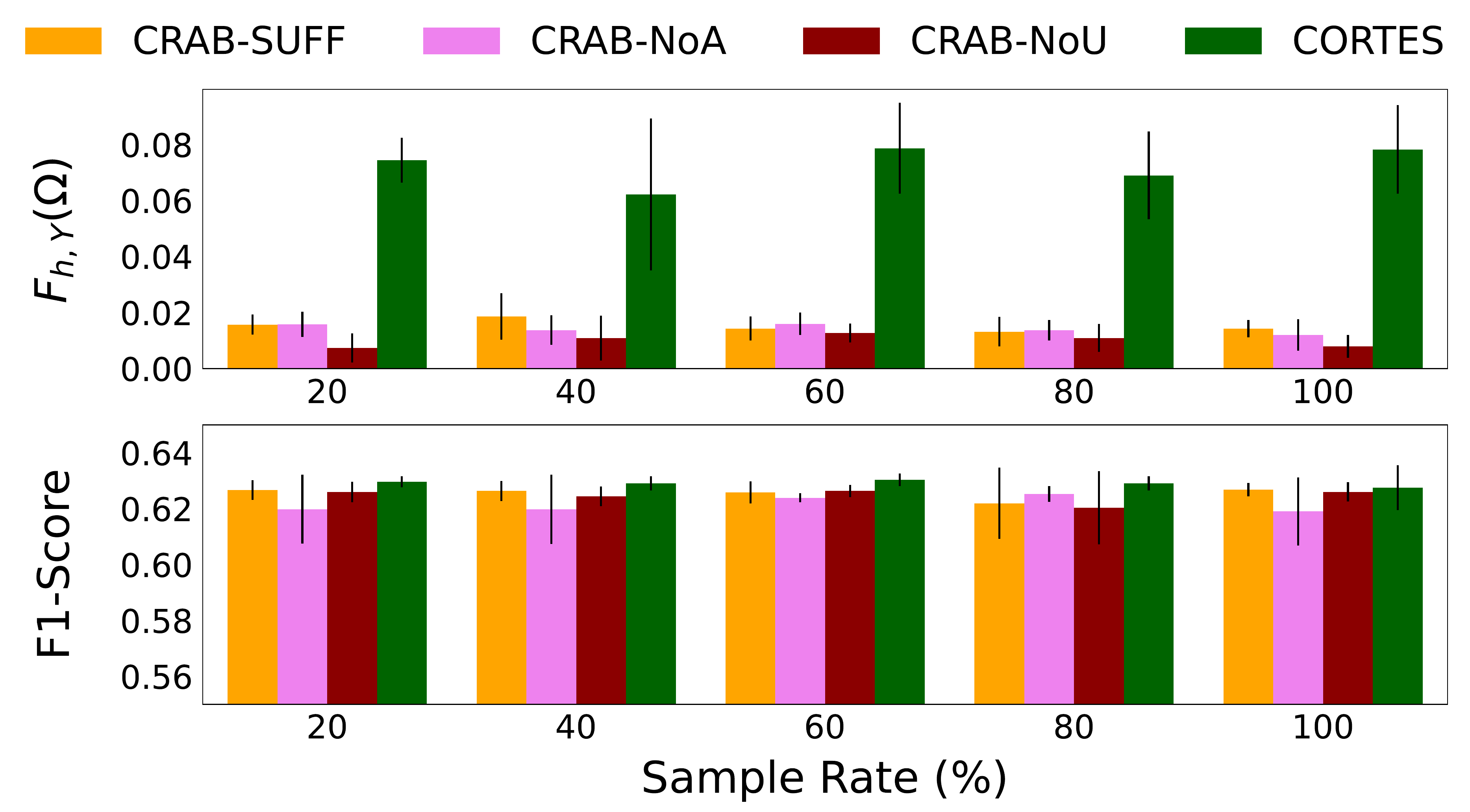}
    \vspace{-4mm}
    \caption{\textmd{The upper plot shows the Equality of Opportunity-Sample Rate; the lower plot shows the F1-Score-Sample Rate.}
    \label{fig:exp:subsample_comparison}}
    \vspace{-4mm}
\end{figure}

\noindent \textbf{Size of external data.}\label{sec:exp:subsample}
We evaluated the impact of varying external data size on the performance of \sys (with $\tau=0.01$) and baselines by using a randomly chosen subset of the unbiased training data as an external data source. Figure~\ref{fig:exp:subsample_comparison} shows that all techniques have similar F1-Scores, however, \sys has a considerably lower $\CondDemoParity{Y}$ compared to \ipw. In fact, \sys achieved the desired level of fairness for all sampling rates and access to external data. On the other hand, \ipw showed the highest standard deviation of unfairness, indicating instability in its performance. As the sample size decreases, the standard deviation of \sys methods increases slightly, indicating a slight decrease in stability as the quality of estimated ratios degrades.

\begin{figure}
    \centering
    \hspace{-0.22cm}
    \includegraphics[scale=0.2]{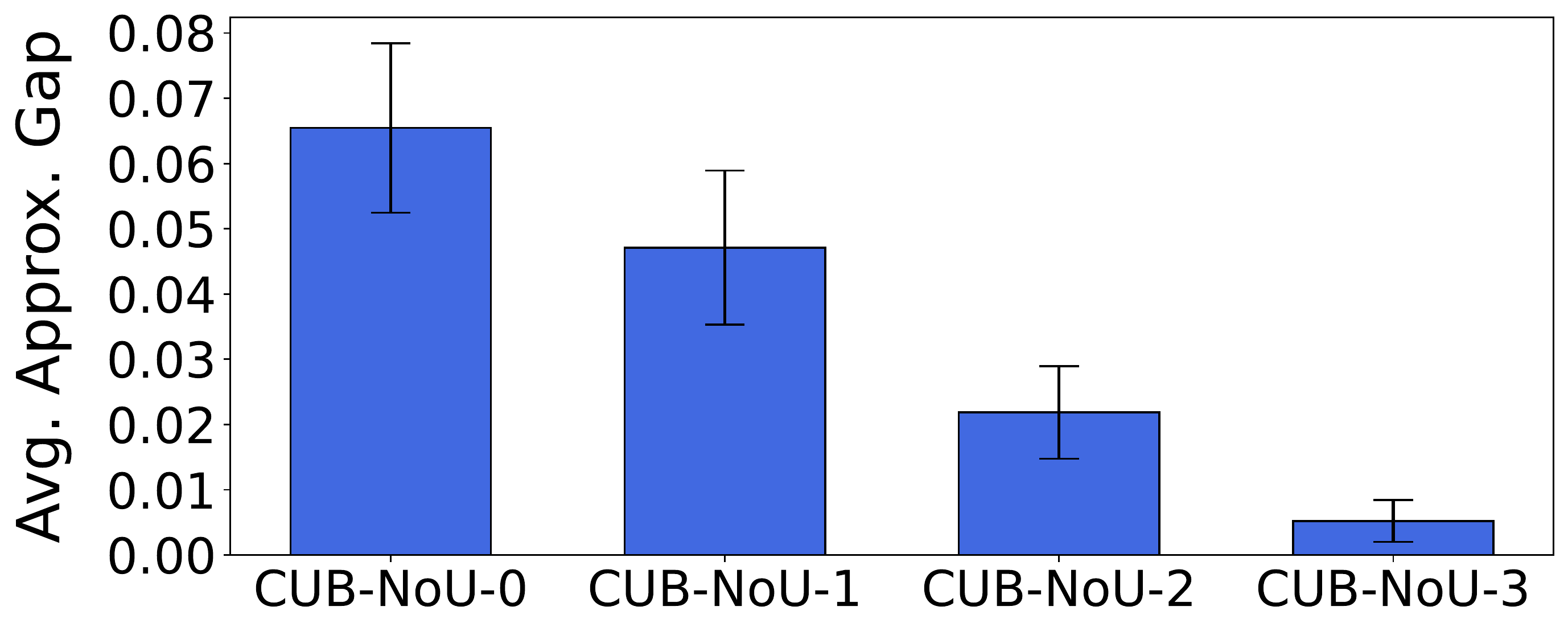}
    \vspace{-4mm}
    \caption{\revc{\textmd{Average Approximation Gap comparison for \sysMissPartU with different levels of external data source.  \CUBMissPartU{n} refers to the CUB obtained in Proposition~\ref{proposition:limited_external_data:dp:max_min_bound} when  $|\Uvar'|=n$.}}
    \label{fig:exp:gap_comparison}}
    \vspace{-0.5cm}
\end{figure}

\revc{
\noindent \textbf{Tightness of CUBs.}
In this experiment, we analyze how the CUBs employed by \sysMissPartU (Proposition~\ref{proposition:limited_external_data:dp:max_min_bound}) differ based on the availability of various levels of external data sources, namely varying sizes of $\Uvar'$.
We trained 100 classifiers with \sysBound on the Adult data with randomly injected selection bias ($|\Uvar|=3$), and compare the average gaps between $\CondDemoParity{Y}$ and CUBs computed with different sizes of $\Uvar'$. 
Our results in 
Figure~\ref{fig:exp:gap_comparison} demonstrate that increasing the size of $\Uvar'$ leads to tighter CUBs. This observation empirically validates the benefits of incorporating more variables into the external data source.
Although \CUBMissPartU{0} has a large approximation gap, it still enables us to train a fair and accurate model, as demonstrated in previous experiments, e.g. Figure~\ref{fig:exp:abs_ext_info_comparison}.
Note that in the absence of any external data source, \sysMissPartU boils down to \sysBound, corresponding to \CUBMissPartU{0}. Likewise, 
when $\Uvar'=\Uvar$, \sysMissPartU is equivalent to \sysComplete, which uses \CUBMissPartU{3}. 
}


%% file: Section_Tex_Files/related_work.tex

\ifextended
Our research relates to previous studies on training fair predictive algorithms, which can be grouped into pre-, post- and in-processing. Pre-processing approaches, most relevant to our study (as surveyed in~\cite{Caton2020FairnessIM}), often depend on the assumption that the training and testing data represent the target population, thus failing to detect discrimination caused by selection bias. Some post-processing approaches that modify the algorithm's decisions based on individuals' sensitive attributes during model deployment are not impacted by selection bias. However, these methods require access to sensitive attribute information during deployment and can only be applied to individuals forming a representative sample of the population, making them highly restrictive and limited in practical applications.
\fi

\reva{Query answering in the presence of selection bias has been studied in the data management community~\cite{orr2020sample,orr2019mosaic}. Also, the effect of incomplete and representation bias in data and its impact on fairness has been studied in databases~\cite{jin2020mithracoverage,shahbazi2023representation}.}
In the context of machine learning, selection bias has also been studied for its impact on models' fairness~\cite{goel2020importance, du2021robust, wang2021analyzing, blum2019recovering, meyer2021certifying}. A study closely related to ours is \cite{goel2020importance}, which examines the problem from a causal viewpoint and discusses how selection bias can cancel out fairness guarantees based on training data. The study investigates different mechanisms corresponding to real-world scenarios of selection bias and determines whether the probability distributions required for different fairness metrics can be recovered in those scenarios. Another study~\cite{wang2021analyzing} analyzes the impact of selection bias on fairness and presents a method to address the issue based on existing methods for learning in the presence of selection bias. Unlike these studies, we rigorously investigate the impact of selection bias on fairness and establish conditions for learning fair ML models in the presence of selection bias with varying degrees of external knowledge about the target population.


Selection bias, which frequently leads to dataset shift, has been extensively researched in the fields of dataset shift and domain adaptation. There are three main categories of dataset shift: covariate shift, prior probability shift, and concept shift, and selection bias can trigger one or more of these shifts~\cite{moreno2012unifying}. Several solutions have been proposed to tackle dataset shifts in classification problems. Reweighting, a common approach, alters the cost of each training point's error based on weights calculated from estimation techniques and a population sample without bias~\cite{cortes2008sample, bareinboim2012controlling, liu2014robust, huang2006correcting, rezaei2020robust}. Nevertheless, past studies have demonstrated that these estimations can result in errors that greatly affect the performance of models in downstream tasks~\cite{cortes2008sample, liu2014robust}, a finding confirmed by our experimental results.
For instance, \cite{maity2021does} determined the necessary and sufficient conditions to enhance the accuracy and fairness of models under covariate shift, while \cite{chen2022fairness} established unfairness limits with an upper boundary on the measure of distribution shift. \cite{mishler2022fair} investigated the impact of concept shift on fairness in performance prediction.
Other research has focused on creating bias reduction methods. \cite{rezaei2020robust} produced a fair model by solving a min-max problem, but it only enforces equalized odds fairness for Logistic Regression models and has limited scalability. \cite{subbaswamy2019preventing,singh2021fairness} proposed causal methods to build fair models based on information about the data collection process, but they concentrate only on covariate shift and work only with addressable graphs. 
Under the assumption that the unbiased data distribution is within the proximity of the biased training distribution, \cite{meyer2021certifying} analyzes the robustness of decision trees; \cite{schumann2019transfer,zhang2021assessing} provide theoretical bounds on fairness in the target distribution using samples from the target distribution; \cite{taskesen2020distributionally} and \cite{wang2021wasserstein} train ML models that are fair on any distribution near the training distribution using distributionally robust optimization. Our work addresses all types of dataset shifts resulting from selection bias.
In comparison, our work addresses any type of dataset shift resulting from selection bias, including combinations of different types, and only requires information about the selection variable's parents in the worst-case scenario, making it more practical compared to other methods that demand knowledge of the target distribution.


\vspace{-0.2cm}
\section{Conclusions}
In this paper, we proposed a novel framework for ensuring fairness in machine learning models trained from biased data. Our framework, inspired by data management principles, presents a method for certifying and ensuring the fairness of predictive models in scenarios where selection bias is present. This framework only requires understanding of the data collection process and can be implemented regardless of the information available from external data sources. Our findings show our framework's success in learning fair models, while accounting for biases in data. It serves as a valuable tool for practitioners striving for fairness in areas often dealing with inherent selection bias.

\newpage

%% file: Section_Tex_Files/discussion.tex
\ignore{\ifextended
\label{sec:discussion}
\subsection{Selection Bias and Dataset Shift}
Selection bias is a major contributor to dataset shift, which occurs when the sample probabilities of the biased population $\popul$ and the target population $\underlyingDist$ are different. There are three main types of dataset shift: covariate shift, prior probability shift, and concept shift, and one or more of them can be caused by selection bias~\cite{moreno2012unifying}. Specifically, a data collection diagram in which the selection occurs based on certain features (e.g., Figure~\ref{figure:structural_conditions:based_on_X2_X4}) can cause covariate shift, where $\pr_{\popul}(y\mid\boldsymbol{x})=\pr_{\underlyingDist}(y\mid\boldsymbol{x})$ and $\pr_{\popul}(\boldsymbol{x})\neq\pr_{\underlyingDist}(\boldsymbol{x})$. On the other hand, a data collection diagram in which the selection depends on both features and the outcome (e.g., Figure~\ref{figure:structural_conditions:based_on_X2_Y}) can cause concept shift, where $\pr_{\popul}(y\mid\boldsymbol{x})\neq\pr_{\underlyingDist}(y\mid\boldsymbol{x})$ and $\pr_{\popul}(\boldsymbol{x})=\pr_{\underlyingDist}(\boldsymbol{x})$. Lastly, a data collection diagram in which the selection is only based on the outcome (e.g., Figure~\ref{figure:structural_conditions:based_on_Y}) can cause prior probability shift, where $\pr_{\popul}(y)\neq\pr_{\underlyingDist}(y)$ and $\pr_{\popul}(\boldsymbol{x}\mid y)=\pr_{\underlyingDist}(\boldsymbol{x}\mid y)$. However, it is important to note that some data collection diagrams can create more complicated cases that do not fit into one of these categories, and therefore, it is insufficient to address selection bias using techniques that only address certain types of dataset shift~\cite{schrouff2022maintaining}.

\subsection{Total Variation Distance}
The total variation distance has been widely used as a measure of the unfairness of an ML model wrt. various fairness notions including conditional statistical parity. It can be used to compute the difference between the output distributions of an ML model with input distributions from two different protected groups.\cite{zhang2018equality,zhang-aaai18}. For example, the output of a biased ML model corresponding to input data from privileged groups has a higher percentage of positive predictions compared to those from protected groups, resulting in a high total variation distance between two output distributions. Further, a fair model that satisfies conditional statistical parity requires the zero total variation distance while adding condition $\Admis=\admis$ to the input distributions for each $\admis\in\Dom{\Admis}$ (each summand in Eq~\eqref{eq:fairnessquery}).
However, there is a line of works that only measure the total variation distance for only single $\admis$\cite{rezaei2020robust}, or for each $\admis\in\Admis$ separately, e.g. showing differences in False Positive Rates ($\Yvar=0$) and True Positive Rates ($\Yvar=1$) separately while enforcing equalized odds\cite{zhang2021omnifair,zhang2018mitigating}. However, our definition of fairness query tries to unify the distances for all $\admis\in\Admis$, which is consistent with the average absolute odds difference (AAOD) that is widely used to measure the violation of equalized odds \cite{moldovan2022algorithmic,bellamy2019ai,kozodoi2022fairness,zhang2021ignorance,stevens2020explainability,madras2018learning}.

\subsection{Implementation Details and Extensions}
\noindent\textbf{Dealing with assumption violation.}
As mentioned while defining Fairness Query in the context of binary classification(Eq~\eqref{eq:binaryfairnessquery}), we assume that $\PrPos_{\underlyingDist }(\classifier(\boldsymbol{x})\mid \privG, \admis)\geq\PrPos_{\underlyingDist }(\classifier(\boldsymbol{x})\mid\protG, \admis)$ holds for each $\admis\in\Admis$. Violation of this assumption will require a minor modification to the established upper bounds of $\CondDemoParityb{\Admis}$ in Proposition \ref{proposition:no_external_data:dp:max_min_bound} and \ref{proposition:limited_external_data:dp:max_min_bound}.
Here we use the Proposition \ref{proposition:no_external_data:dp:max_min_bound} as an example. When the assumption does not hold for some $\admis\in\Admis$, we can do the reverse for the corresponding summand in Eq~\eqref{eq:bound:no-external}: using the upper bound of $\PrPos_{\popul}(h(\boldsymbol{x})\mid\protG, \admis)$ minus the lower bound of $\PrPos_{\popul}(h(\boldsymbol{x})\mid\privG, \admis)$.

\noindent\textbf{Adaptation to Multi-class Classification and Regression.} The fairness query defined in Eq~\eqref{eq:fairnessquery} allows $\Dom{\Yvar}$ to include multiple outcome values, which corresponds to multi-class classification tasks. Hence the proposed theorems can be easily adapted to multi-class classification by adding a summation on Y values. Similarly, the most straightforward way to adapt \sys to a regression problem is to discretize the outcome \Yvar\ and convert this to a multi-class classification problem\cite {agarwal2019fair}.

\noindent\textbf{Adaptation to Multiple Protected Groups.} When there exist multi-valued protected attributes, or multiple protected attributes, the training data can be divided into more than 2 protected groups. For example, when we have 2 binary protected attributes $\ProtectedAttr_1$ and $\ProtectedAttr_2$, there are 4 protected groups divided by conditions: ($\ProtectedAttr_1=0,\ProtectedAttr_2=0$), ($\ProtectedAttr_1=1,\ProtectedAttr_2=0$), ($\ProtectedAttr_1=0,\ProtectedAttr_2=1$), ($\ProtectedAttr_1=1,\ProtectedAttr_2=1$). In this case, we can apply theorems on each pair of protected groups, thereby each pair corresponds to one lowest upper bound of unfairness. Then the average of all upper bounds can be taken as the final penalty term in Eq~\eqref{eq:reqopt}. This is also consistent with the fairness notion of average statistical imparity used in \cite{kang2021multifair} that takes the average of statistical imparities computed between all pairs of protected groups.

\subsection{Limitations}
\noindent\textbf{Access to data collection diagrams.}
Although neither the entire data collection diagram nor the selection probabilities are required, \sys still needs at least the parents of the selection variable in order to ensure fairness in the target population. This is a weaker assumption than \cite{subbaswamy2019preventing,singh2021fairness} that use the entire causal diagram for bias mitigation in the context of dataset shift. However, even when we have no direct access to the underlying data collection diagram, but have samples from the unbiased distribution (a popular setting in domain adaptation literature, see Section~\ref{sec:discussion}), we can construct the data collection diagram via causal discovery techniques\cite{subbaswamy2020spec,sun2021recovering}. These techniques allow us to find the selection variable $\SelectionVar$ and its dependency with other variables including $\pa(\SelectionVar)$, based on which \sys can be applied.

\noindent\textbf{Discretization.}
In this work, even the conditional independence encoded in the data collection diagram can be leveraged and applied to arbitrary form of data, \sys still relies on discretized data to compute the consistent upper bound of the fairness query answer. This is because the summation over the domain of discrete variables cannot be directly translated to integrals over the domain of continuous variables in practice, unless we are fully aware of the data distribution, e.g. know the sample probability for arbitrarily possible data point. This might be realized via generative models by mimicking the original data collection process, but we defer it to future works.
\fi}